\pdfoutput=1

\documentclass[pmlr,twocolumn,10pt]{jmlr} 

\usepackage{booktabs}
\usepackage{multirow}
\usepackage{graphicx}
\usepackage{siunitx}

\usepackage{enumitem}

\newcommand{\equal}[1]{{\hypersetup{linkcolor=black}\thanks{#1}}}

\theorembodyfont{\upshape}
\theoremheaderfont{\scshape}
\theorempostheader{:}
\theoremsep{\newline}

\jmlrvolume{LEAVE UNSET}
\jmlryear{2022}
\jmlrsubmitted{LEAVE UNSET}
\jmlrpublished{LEAVE UNSET}
\jmlrworkshop{Conference on Health, Inference, and Learning (CHIL) 2022} 

\title[MS disability prediction]{Disability prediction in multiple sclerosis using performance outcome measures and demographic data}

\author{%
 \Name{Subhrajit Roy}\equal{These authors contributed equally}
 \Email{subhrajitroy@google.com}\\
 \Name{Diana Mincu}\footnotemark[1] \Email{dmincu@google.com} \\
 \Name{Lev Proleev} \Email{levp@google.com} \\
 \Name{Negar Rostamzadeh} \Email{nrostamzadeh@google.com}\\
 \Name{Chintan Ghate} \Email{chintanghate@google.com}\\
 \Name{Natalie Harris} \Email{natalieharris@google.com}\\
 \Name{Christina Chen} \Email{christinium@google.com}\\
 \Name{Jessica Schrouff} \Email{schrouff@google.com}\\
 \addr Google Research \\
 \Name{Nenad Tomašev} \Email{nenadt@deepmind.com} \\ \addr DeepMind\\
 \Name{Fletcher Lee Hartsell} \Email{fletcher.hartsell@duke.edu} \\ \addr Duke University Health System\\
 \Name{Katherine Heller} \Email{kheller@google.com} \\ \addr Google Research\\
 \Name{for MSOAC}\equal{Data used in the preparation of this article were obtained from the Multiple Sclerosis Outcome Assessments Consortium (MSOAC).  As such, the investigators within MSOAC contributed to the design and implementation of the MSOAC Placebo database and/or provided placebo data, but did not participate in the analysis of the data or the writing of this report.}\\
}

\begin{document}

\maketitle

\begin{abstract}
Literature on machine learning for multiple sclerosis has primarily focused on the use of neuroimaging data such as magnetic resonance imaging and clinical laboratory tests for disease identification. However, studies have shown that these modalities are not consistent with disease activity such as symptoms or disease progression. Furthermore, the cost of collecting data from these modalities is high, leading to scarce evaluations. In this work, we used multi-dimensional, affordable, physical and smartphone-based performance outcome measures (POM) in conjunction with demographic data to predict multiple sclerosis disease progression. We performed a rigorous benchmarking exercise on two datasets and present results across 13 clinically actionable prediction endpoints and 6 machine learning models. To the best of our knowledge, our results are the first to show that it is possible to predict disease progression using POMs and demographic data in the context of both clinical trials and smartphone-based studies by using two datasets. Moreover, we investigate our models to understand the impact of different POMs and demographics on model performance through feature ablation studies. We also show that model performance is similar across different demographic subgroups (based on age and sex). To enable this work, we developed an end-to-end reusable pre-processing and machine learning framework which allows quicker experimentation over disparate MS datasets.
\end{abstract}

\section{Data and Code Availability}
This paper uses two publicly available datasets: Multiple Sclerosis Outcome Assessments Consortium (MSOAC) (\cite{Rudick2014}, \url{https://c-path.org/programs/msoac/}) and Floodlight (\cite{BakerFloodlight}, \url{https://floodlightopen.com/en-US/for-scientists}). While our code is not available at this time, we plan to open-source it in the future.

\section{Introduction}
\label{sec:intro}

Multiple sclerosis (MS) is a neurological disease that affects around 2.8 million people worldwide and is the leading cause of non-traumatic disability in young adults \citep{atlas_of_ms_2020}. The primary goal of a clinician treating MS is to manage disease activity and reduce the risk of disability. As such, the ability to accurately predict MS disease progression has the potential to guide therapy and may inform decisions about the most effective care. While machine learning (ML) models have been developed for predicting disease progression in MS \citep{Pinto2020,Zhao2017,Rodriguez2012,Seccia2020,Tommasin2021}, these approaches primarily rely on using clinically-acquired information such as magnetic resonance imaging (MRI) \citep{Zhao2017}, clinical laboratory tests or clinical history \citep{Seccia2020}. A lack of association between disease activity and these modalities has previously been identified \citep{Whitaker1995}, which in turn led to the development of multi-dimensional performance outcome measures (POMs) such as Multiple sclerosis functional composite (MSFC) scores to accurately track MS disease progression \citep{MSFC_paper}. POMs are time-stamped responses collected from MS subjects either through assessment tests or questionnaires, which are used to track disease progression. These include tests to quantify walking ability, balance, cognition, and dexterity – physiological functions that are adversely affected by MS. The frequency of data collection may vary with intervals ranging from a day to multiple months. In addition, they also reduce costs related to personnel, equipment, space, and time requirements compared to neuroimaging or clinical laboratory tests. POMs have also been used alongside neuroimaging-derived data for predicting disability in MS \citep{Law2019}. Moreover, while POMs and demographic data have been used to diagnose MS \citep{Patrick2021}, these have not been used for continually predicting MS disability progression.

In this work, we investigate the possibility of using POMs (physical or electronic) and demographic data for predicting disease progression (in particular disability scores), in MS subjects, in both a clinical and at-home setting. We proof-test this idea using two openly accessible MS datasets: MSOAC \citep{MSOAC_paper} and Floodlight \citep{BakerFloodlight}. Our contributions are as follows:

\begin{enumerate}
[noitemsep,nolistsep]
    \item \textit{Novel ML Health solution:} We present for the first time (to the best of our knowledge) that the conjunction of POMs and demographic data can be used to successfully and continually predict long- and short-term MS disease progression for clinical and smartphone-based datasets.
    \item \textit{Additional analysis:} We show that model performance is similar across different demographic subgroups (based on age and sex), and perform multiple feature ablations to understand the contributions of different POMS and demographics to the predictions.
    \item \textit{Reliability and scalability:} We present a reusable end-to-end pre-processing and machine learning modelling framework that enables benchmarking on different MS datasets. Our proposed framework focuses on reliable dataset ingestion through a common format, scalable label creation and metrics computation.
\end{enumerate}

We envision that our work will not only serve as a first step towards development of machine learning models for monitoring MS, but also spur more ML research in this application area.

\section{Methods}
\label{sec:methods}

\subsection{Data description}
\label{subsec:data_description}

\begin{table*}[h]
\centering
\caption{Comparison of the MSOAC and Floodlight datasets.}
\label{tab:dataset_comparison}
\resizebox{0.8\textwidth}{!}{%
    \begin{tabular}{|l|l|l|}
    \hline
     & \textbf{MSOAC} & \textbf{Floodlight}  \\ \hline
    Modality & Clinical visit & Smartphone \\ \hline
    Cohort & MS subjects only & MS subjects + control \\ \hline
    Frequency of assessment  & 3-monthly & Continuous \\ \hline
    Test type & \begin{tabular}[c]{@{}l@{}}Physical Multiple Sclerosis \\ Functional Composite (MSFC) scores\end{tabular} & Smartphone-based \\ \hline
    Clinician annotation & \begin{tabular}[c]{@{}l@{}}Expanded Disability Status Scale (EDSS)\end{tabular} & None \\ \hline
    Number of patients & 2,465 & 2,339 \\ \hline
    \end{tabular}%
}
\end{table*}
We looked at two datasets for benchmarking, one recorded in a clinical trial setting (MSOAC), and one from a mobile app in a clinically unsupervised manner (Floodlight). The MSOAC dataset records POMs from physical MSFC tests which were performed by the subjects in-clinic as a part of clinical trials, while the Floodlight dataset records outcome measures collected via an electronic equivalent of MSFC tests taken by the subjects on a smartphone. Both have been previously used for machine learning explorations \citep{Schwab20, walsh2020generating}. A comparison of the two datasets can be seen in Table \ref{tab:dataset_comparison} and a set of statistics on the data can be found in Appendix \ref{prediction_task_stats}.

\noindent \textbf{MSOAC Placebo Database:}
The Multiple Sclerosis Outcome Assessments Consortium (MSOAC, \citep{Rudick2014}) was launched in 2012 to collect, standardize, and analyze data about MS. To that end, their Placebo Database collects data from the placebo arms of 9 different clinical trials \citep{MSOAC_paper} with 2,465 individual patient records.

It contains information on: demographics, medical history, POMs (e.g. timed walk test, dexterity tests, auditory and visual acuity tests), patient reported outcome measures (e.g. health survey), relapse information and the MS sub-type, and clinically vetted measurements such as the Expanded Disability Status Scale (EDSS) \citep{Kurtzke1444}.

\noindent \textbf{Floodlight:}
Floodlight is a mobile app developed by Roche and Genentech \citep{BakerFloodlight} designed to combat the infrequent measurements observed during clinical visits and allow healthcare professionals to have a greater understanding of the disease. It contains a set of active tests that measure brain function (daily mood question, symbol matching), hand function (draw a shape, on-screen pinching) and mobility (timed two minute walk, balance, u-turn). Unlike MSOAC, it does not contain any clinically or expert defined labels.

The app is still active at the time of writing and the number of enrolled patients is constantly growing. The data snapshot we use was taken on the 15th of June 2021 and has a total of 2,339 subjects, including both MS patients ($n=1,236$) and control subjects ($n=1,103$).

\subsection{Data processing}
\label{subsec:data_processing}

\begin{figure*}
  \includegraphics[width=\linewidth]{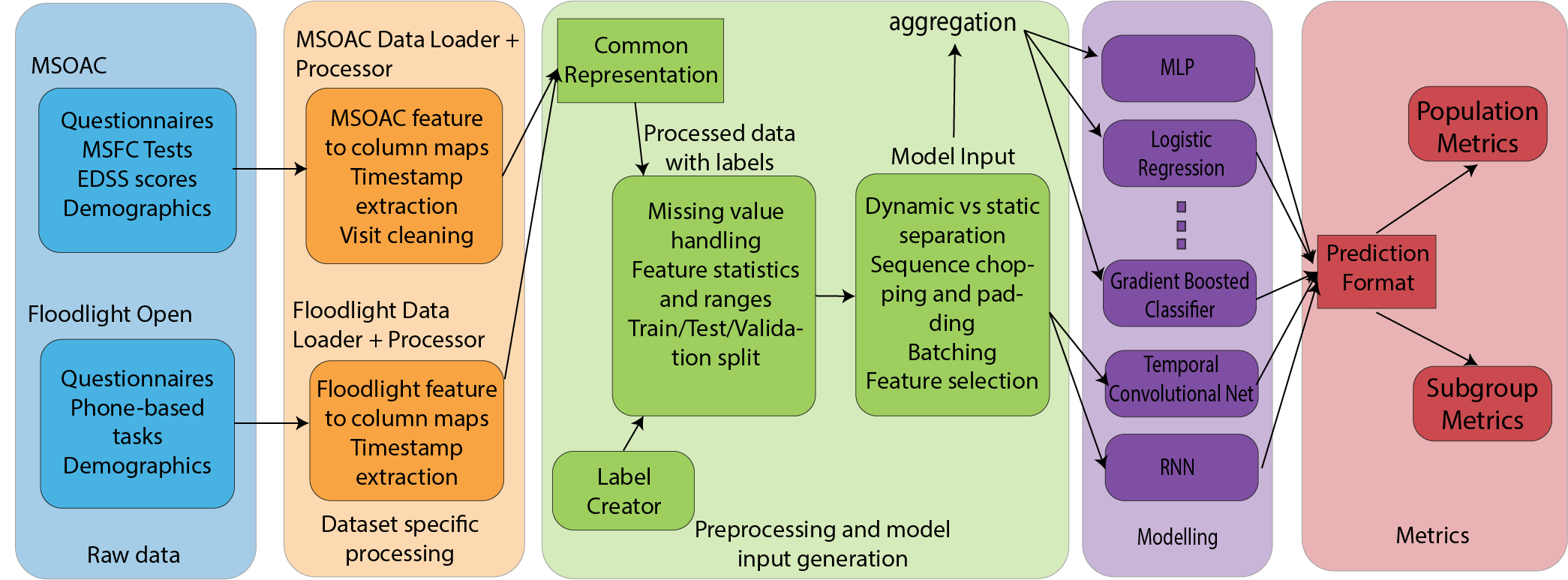}
  \caption{Dataset processing pipeline with common downstream modelling infrastructure}
  \label{Processing_pipeline}
\end{figure*}

Since the type and structure of the data contained in the two datasets are quite different, typically all data processing would be done individually for each dataset. This can lead to slight differences in the resulting model input, making direct comparison between datasets difficult. To avoid this pitfall, we have devised a general data processing, modelling and evaluation pipeline (shown in Figure \ref{Processing_pipeline}) which enables us to reuse a number of downstream components and do a reliable cross-dataset comparison.

\subsubsection{Pipeline overview}

The raw data is taken through a set of processing steps into a common representation called Subject, inspired from \citep{Tomasev2021}. Once both the clinical dataset and the smartphone-based dataset are transformed into the Subject representation, a Label Creator runs on all processed data and enriches it with labels (see Sec. \ref{subsec:tasks} for a description of the tasks), after which it gets transformed into model input. The labels can be dataset-specific, or common across multiple different datasets. The proposed pipeline is feature agnostic i.e. there is no specific pre-processing in one dataset or the other.

After training, a prediction format (Table A\ref{prediction_representation}, Appendix) is used to save all model output. This in turn gets fed into the metrics pipeline which can provide results at both population and subgroup levels.

\subsubsection{Common format}

A full description of each field present in this representation can be found in Appendix \ref{subject_proto_details}. At a high level, each subject has some information that is constant across time (static) such as the medical history or the subject's sex \footnote{In MSOAC this is clinician reported, while in Floodlight it is self reported.}, but also timestamp-based information (dynamic) encompassing all medical events - either outpatient or inpatient. Multiple such Events can be grouped into an Episode, but an individual Event can also form an Episode on its own.

The part that enables each dataset to be processed individually is the concept of Resources. These can be functional tests, questionnaires, medications or more, depending on the types of data available in each dataset. Each medical event has a set of resources associated with it, and since the types of resources depend on the dataset, this leaves a common overall structure while still allowing for variability.

To better illustrate this structure we can look at what this means for our two datasets. In the case of MSOAC, an Episode corresponds to a visit to the clinic, including all the tests performed. The data available at each visit consists of functional tests, questionnaires and medications. Each of these is represented as a Resource. Given that MSOAC does not provide timestamps for these resources, they are considered as part of the same Event. For Floodlight, the dataset contains less but more frequent and timestamped measurements. In this case, we define an Episode as 24 hours of data, and each Event corresponds to a new POM. The two types of resources available are functional tests and questionnaires.

We emphasise that the datasets are not merged, but separately converted into the common Subject representation. Our goal is to demonstrate that ML modelling for disability prediction in MS is possible for two disparate datasets using the same modelling framework.

\subsection{Prediction tasks}
\label{subsec:tasks}
The disability prediction tasks are selected such that they are \textit{clinically actionable} in the context of the particular dataset they are defined in.

\subsubsection{MSOAC}
\paragraph{EDSS:}\label{edss_derived_labels} For MSOAC our primary prediction endpoints are derived from clinician-annotated EDSS scores which is a commonly used measure of long-term MS disability \citep{Kurtzke_edss_paper}. The EDSS scale ranges from 0 to 10, in 0.5 unit increments. EDSS scores can be divided into distinct severity categories: 0-1 for no disability, 1.5-2.5 for mild disability, 3-4.5 for moderate disability, and 5-10 for severe disability. In this paper, we consider both the prediction of raw EDSS scores ($EDSS_{mean}$) and more clinically insightful tasks such as the severity of EDSS scores and whether it crosses a certain disability risk-threshold. Detecting a change in EDSS severity could signal to a clinician the need to change the medication a patient is on, or be used to check whether a treatment is effective.

All tasks are implemented as continuous predictions, triggered at every visit. Fixed prediction horizons are chosen for each task based on expert clinical input on the window of actionability (Table \ref{tab:msoac_full_feature_results}). These are 0 - 6 months, 6 - 12 months, 12 - 18 months and 18 - 24 months.

\subsubsection{Floodlight}
\paragraph{Disability scores:}\label{floodlight_disability_scores}

Floodlight does not contain any expert annotations, so we developed a score that closely mimics EDSS. EDSS is divided into multiple components measuring different kinds of disability: neural function, ambulatory, and walking. We categorize the assessment tests present in the Floodlight dataset into the above categories and perform a weighted combination of the tests in each category to develop three individual disability scores for separate functional systems. We also compute an overall disability score by taking the average of the individual functional scores. Given no literature exists on how to define disability scores from smartphone-based assessment tests, we rely on expert input and expect the score to be a close proxy of EDSS. These tasks are formulated as regression tasks since smartphone-based tests are relatively new and hence severity categories are not defined in literature. For the purpose of this work, we assume that continuous predictions of this derived disability score provides insights on the progression of the disease. While the EDSS-derived labels in Sec. \ref{edss_derived_labels} track long-term changes in disability, the higher frequency recordings of Floodlight enables the prediction of short-term changes.  

Similar to the section above, all tasks are posed as continuous predictions, triggered after a new POM is available. Prediction horizons for these tasks are smaller than for MSOAC:  0 - 1 weeks, 1 - 2 weeks and 2 - 4 weeks (Table \ref{tab:floodlight_full_feature_results}). This is because (a) frequent measurements are possible via a smartphone and hence short-term changes in disability trends can be predicted and  (b) most Floodlight users stop using the app after a certain point, so long-term data is not available.

\paragraph{Smartphone-based diagnosis of MS:}
\label{smartphone_ms_diagnosis}

While predicting disability progression in MS is beneficial in both a clinical and at-home context, earlier diagnosis and treatment of MS is considered the best path to fighting it \citep{pmid15253684}. Multiple studies have shown a delay between symptom development, to first medical visit and then finally to diagnosis, with an average of 1-2 years between symptom onset and diagnosis \citep{Ghiasian2021, Fernandez2010}. We believe that the usage of smartphones can enable large-scale diagnosis of MS in a more timely manner.

To test this hypothesis, we use the Floodlight dataset to predict whether a subject has MS or not. This is not possible in MSOAC as we do not have control subjects. The problem statement is formulated as follows: we are given $N$ POMs in total for each subject since the start of data collection, along with the self-reported ground truth on whether or not they have MS. The models have to predict whether the set of $N$ tests belong to a subject with MS. We vary N as = {5, 10, 20, 30, 40, 50, 60, 70, 80, 90, 100}. Note that it is desirable to use less tests for making this prediction, not only to enable earlier access to clinical care, but also to tackle the adherence problem which plagues smartphone-based health studies \citep{Patoz2021, Pathiravasan2021}.

\subsubsection{Cross-dataset}

\paragraph{Disability progression:}
\label{subsec:disab_prog_labels}

To investigate differences in signal between the two datasets, we also define a common label across both MSOAC and Floodlight (Figure A\ref{disability_progression_labels}, Appendix \ref{appendix:disability_progression}). It is focused on forecasting disability by predicting substantial deviations of questionnaire and functional test values. This is because: (a) EDSS or other aggregated disability scores are combined across functional systems and (b) EDSS has been criticised to be focused more on mobility and less on cognitive abilities or dexterity \citep{Meyer-Moock2014}. Successfully predicting the deviation of individual functional tests can potentially be more informative for clinicians in understanding which functional systems of a subject are likely to contribute to a subject's future disability. We define the disability progression labels as a change (greater/lesser) of 20\% \citep{Goldmane1921} from a baseline, where the baseline is updated as time progresses, for each subject and each feature. This led to a three-class classification problem where each timestamp was annotated with one of the three labels: disability unchanged, improved, or worsened.

\subsection{Features and Missingness}
\label{subsec:feature_selection}

We define a set of input features for each task to prevent any label leakage during the training and testing of the model. For MSOAC, we eliminate the EDSS score feature for EDSS-derived targets. For both datasets we only use the questionnaires, functional tests, and patient characteristics such as age, sex, weight, height (dataset dependent, where available). In total, there are 92 distinct features for MSOAC (POMs are multi-component) and 24 for Floodlight.  

\begin{figure}[!ht]
\begin{center}
\centerline{\includegraphics[width=1\columnwidth]{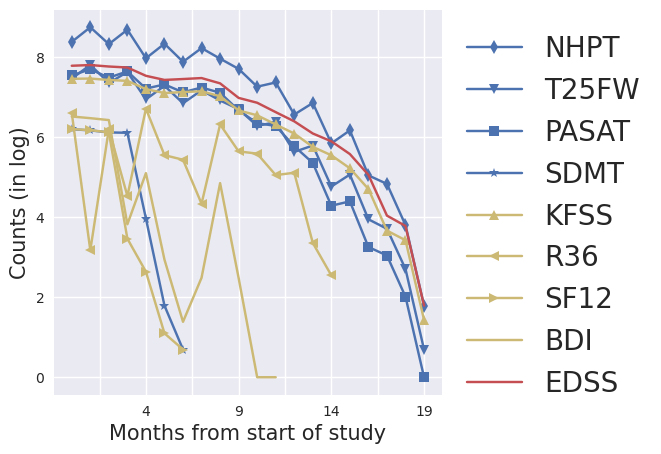}}
\caption{Feature sparsity in MSOAC}
\label{feature_sparsity_msoac}
\end{center}
\end{figure}

To evaluate feature sparsity in our datasets, we look at population level counts for each feature and a given time bucket. For MSOAC we use buckets of 1 month from the start of the study, while for Floodlight we use 1 week intervals from the time the subject joined the app. Based on figure \ref{feature_sparsity_msoac} we see that in MSOAC functional tests (Nine Hole Peg Test (NHPT), Timed 25-Foot Walk (T25FW), Paced Auditory Serial Addition Test (PASAT), Symbol Digit Modalities Test (SDMT)) have a much higher count than questionnaires (Kurtzke Functional Systems Scores (KFSS), RAND-36 Item Health Survey (RAND-36), 12-Item Short Form Survey (SF-12), Beck Depression Inventory (BDI)) and are present until much later in the study as well. For Floodlight (Appendix \ref{appendix:feature_stats}, Figure A\ref{feature_sparsity_floodlight}) we observe a similar downward trend as time progresses, but in this case the functional tests and questionnaires behave in a similar manner. This can be explained by the simplicity of the questionnaires in this dataset, since only one mood related question is available. 

Missing values at various timestamps were replaced by 0 in the case of numerical features, or empty string for text-based features.

\begin{figure*}[!ht]
  \centering
  \includegraphics[width=0.8\linewidth]{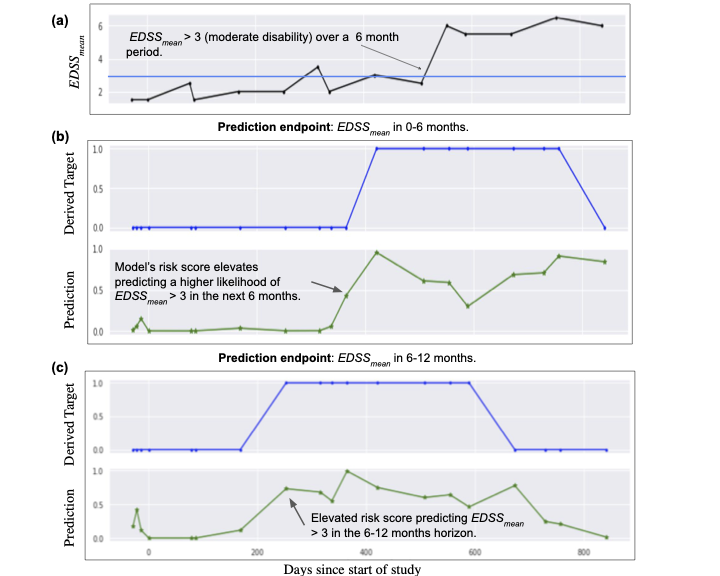}
  \caption{Illustrative example of the trained model successfully predicting that a female subject of age 29 is going to have moderate disability ($EDSS_{mean} > 3$) in the (b) 0-6 and (c) 6-12 months horizon.}
  \label{edss_pred_figure}
\end{figure*}

\subsection{Models}
\label{subsec:models}
For benchmarking purposes, we choose a few popular baseline models from Scikit-Learn (Logistic Regression, Linear Regression, Gradient Boosted Classifier, Gradient Boosted Regressor), a non-sequential (Multi-layer Perceptron), and a sequential deep neural network (Temporal Convolutional Neural Networks (TCN) \citep{bai2018empirical}.

For predictions, the models use information across a specified window before the prediction timepoint. While for Scikit-Learn models information is aggregated across the window by taking the mean, TCN processes them sequentially and hence retains the temporal information. 

We report the area under the precision-recall curve (AU PRC) for the classification tasks since most prediction problems are imbalanced (see Appendix \ref{prediction_task_stats}, Table \ref{tab:label_stats}) and R-MSE for regression tasks. We perform 10-fold cross-validation for each model and report the mean and standard deviation of AU PRC / R-MSE. For TCN we use the Adam optimizer \citep{Kingma2015}. We performed a hyperparameter search for each model to find the optimal hyperparameters, and the search space is reported in Appendix \ref{appendix:hyperparameters}.

\section{Experiments and results}
\label{sec:results}

\subsection{Performance on the full feature set}

\subsubsection{MSOAC}

\noindent \textbf{Illustrative example:}
Figure \ref{edss_pred_figure} shows an example usage of our predictive models for the MSOAC dataset. The model is trained to predict whether the patient will transition to a state of moderate disability in the next 0-6 months and 6-12 months interval. Updated risk estimates of future disease worsening are made for every clinic visit throughout the course of the clinical study. Identifying an increased risk of decline sufficiently well in advance can enable early preventative action \citep{long_pred_pred_disab}. This is possible even when clinicians may not be monitoring a patient or actively intervening.

\begin{figure}[!ht]
  \centering
  \includegraphics[width=\linewidth]{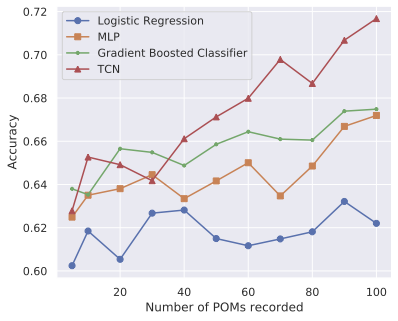}
  \caption{Comparison of model performance for diagnosing MS on Floodlight (smartphone-based dataset). }
  \label{ms_vs_non_ms}
\end{figure}

\noindent \textbf{Performance on EDSS-derived labels:}
Table \ref{tab:msoac_full_feature_results} summarizes model performance on the labels (see Sec.\ \ref{edss_derived_labels}) derived from EDSS scores. We show the mean and standard deviation of the metrics across different folds. We observe that TCN consistently outperforms other ML algorithms by achieving superior performance in all regression and classification tasks. We also see that although the dataset is composed of data from 9 different clinical trials, the models still manage to learn meaningful representations, as demonstrated by the results. For MSOAC, we observe a general trend that, as the prediction interval slides into the future, the tasks progressively get more difficult, leading to a reduction in model performance. This phenomenon has also been shown in other sequential healthcare prediction problems \citep{Tomasev2021, Nestor2019}.

\begin{table*}[!ht]
\centering
\caption{Performance (and standard deviation) obtained by machine learning models on a diverse set of prediction tasks for the MSOAC dataset. Best performance is reported in bold.}
\label{tab:msoac_full_feature_results}
\resizebox{1.0\textwidth}{!}{%
\begin{tabular}{|l|l|l|l|l|l|l|l|l|l|}
\hline
\textbf{Dataset} &
  \begin{tabular}[c]{@{}l@{}}\textbf{Prediction}\\ \textbf{tasks}\end{tabular} &
  \begin{tabular}[c]{@{}l@{}}\textbf{Prediction}\\ \textbf{Window}\end{tabular} &
  \textbf{Metric} &
  \begin{tabular}[c]{@{}l@{}}\textbf{Logistic}\\ \textbf{Regression}\end{tabular} &
  \begin{tabular}[c]{@{}l@{}}\textbf{Linear}\\ \textbf{Regression}\end{tabular} &
  \textbf{MLP} &
  \begin{tabular}[c]{@{}l@{}}\textbf{Gradient}\\ \textbf{Boosted}\\ \textbf{Classifier}\end{tabular} &
  \begin{tabular}[c]{@{}l@{}}\textbf{Gradient}\\ \textbf{Boosted}\\ \textbf{Regressor}\end{tabular} &
  \begin{tabular}[c]{@{}l@{}}\textbf{TCN}\end{tabular} \\ \hline
\multirow{4}{*}{MSOAC} &
  \multirow{4}{*}{$EDSS_{mean}$} &
  0 - 6 mo &
  \multirow{4}{*}{R-MSE} &
  - &
  \begin{tabular}[c]{@{}l@{}}1.929 (0.098)\end{tabular} &
  - &
  - &
  \begin{tabular}[c]{@{}l@{}}1.700 (0.045)\end{tabular} &
  \textbf{\begin{tabular}[c]{@{}l@{}}1.264 (0.055)\end{tabular}} \\ \cline{3-3} \cline{5-10} 
 &
   &
  6 - 12 mo &
   &
  - &
  \begin{tabular}[c]{@{}l@{}}2.114 (0.111)\end{tabular} &
  - &
  - &
  \begin{tabular}[c]{@{}l@{}}1.901 (0.057)\end{tabular} &
  \textbf{\begin{tabular}[c]{@{}l@{}}1.650 (0.067)\end{tabular}} \\ \cline{3-3} \cline{5-10} 
 &
   &
  12 - 18 mo &
   &
  - &
  \begin{tabular}[c]{@{}l@{}}2.186 (0.115)\end{tabular} &
  - &
  - &
  \begin{tabular}[c]{@{}l@{}}1.892 (0.065)\end{tabular} &
  \textbf{\begin{tabular}[c]{@{}l@{}}1.725 (0.074)\end{tabular}} \\ \cline{3-3} \cline{5-10} 
 &
   &
  18 - 24 mo &
   &
  - &
  \begin{tabular}[c]{@{}l@{}}2.068 (0.133)\end{tabular} &
  - &
  - &
  \begin{tabular}[c]{@{}l@{}}1.748 (0.062)\end{tabular} &
  \textbf{\begin{tabular}[c]{@{}l@{}}1.666 (0.128)\end{tabular}} \\ \cline{2-10} 
 &
  \multirow{4}{*}{\begin{tabular}[c]{@{}l@{}}$EDSS_{mean}$\\ $ > 3$ (Moderate\\ disability)\end{tabular}} &
  0 - 6 mo &
  \multirow{4}{*}{AU PRC} &
  \begin{tabular}[c]{@{}l@{}}0.803 (0.012)\end{tabular} &
  - &
  \begin{tabular}[c]{@{}l@{}}0.826 (0.015)\end{tabular} &
  \begin{tabular}[c]{@{}l@{}}0.843 (0.017)\end{tabular} &
  - &
  \textbf{\begin{tabular}[c]{@{}l@{}}0.909 (0.014)\end{tabular}} \\ \cline{3-3} \cline{5-10} 
 &
   &
  6 - 12 mo &
   &
  \begin{tabular}[c]{@{}l@{}}0.707 (0.014)\end{tabular} &
  - &
  \begin{tabular}[c]{@{}l@{}}0.731 (0.019)\end{tabular} &
  \begin{tabular}[c]{@{}l@{}}0.756 (0.019)\end{tabular} &
  - &
  \textbf{\begin{tabular}[c]{@{}l@{}}0.82 (0.027)\end{tabular}} \\ \cline{3-3} \cline{5-10} 
 &
   &
  12 - 18 mo &
   &
  \begin{tabular}[c]{@{}l@{}}0.605 (0.025)\end{tabular} &
  - &
  \begin{tabular}[c]{@{}l@{}}0.664 (0.036)\end{tabular} &
  \begin{tabular}[c]{@{}l@{}}0.706 (0.027)\end{tabular} &
  - &
 \textbf{\begin{tabular}[c]{@{}l@{}}0.768 (0.031)\end{tabular}} \\ \cline{3-3} \cline{5-10} 
 &
   &
  18 - 24 mo &
   &
  \begin{tabular}[c]{@{}l@{}}0.502 (0.036)\end{tabular} &
  - &
  \begin{tabular}[c]{@{}l@{}}0.594 (0.038)\end{tabular} &
  \begin{tabular}[c]{@{}l@{}}0.641 (0.038)\end{tabular} &
  - &
  \textbf{\begin{tabular}[c]{@{}l@{}}0.703 (0.038)\end{tabular}} \\ \cline{2-10} 
 &
  \multirow{4}{*}{\begin{tabular}[c]{@{}l@{}}$EDSS_{mean}$\\ $ > 5$ (Severe\\ disability)\end{tabular}} &
  0 - 6 mo &
  \multirow{4}{*}{AU PRC} &
  \begin{tabular}[c]{@{}l@{}}0.695 (0.026)\end{tabular} &
  - &
  \begin{tabular}[c]{@{}l@{}}0.727 (0.026)\end{tabular} &
  \begin{tabular}[c]{@{}l@{}}0.785 (0.025)\end{tabular} &
  - &
  \textbf{\begin{tabular}[c]{@{}l@{}}0.848 (0.035)\end{tabular}} \\ \cline{3-3} \cline{5-10} 
 &
   &
  6 - 12 mo &
   &
  \begin{tabular}[c]{@{}l@{}}0.576 (0.028)\end{tabular} &
  - &
  \begin{tabular}[c]{@{}l@{}}0.597 (0.027)\end{tabular} &
  \begin{tabular}[c]{@{}l@{}}0.676 (0.032)\end{tabular} &
  - &
  \textbf{\begin{tabular}[c]{@{}l@{}}0.722 (0.039)\end{tabular}} \\ \cline{3-3} \cline{5-10} 
 &
   &
  12 - 18 mo &
   &
  \begin{tabular}[c]{@{}l@{}}0.457 (0.035)\end{tabular} &
  - &
  \begin{tabular}[c]{@{}l@{}}0.504 (0.032)\end{tabular} &
  \begin{tabular}[c]{@{}l@{}}0.594 (0.034)\end{tabular} &
  - &
  \textbf{\begin{tabular}[c]{@{}l@{}}0.669 (0.037)\end{tabular}} \\ \cline{3-3} \cline{5-10} 
 &
   &
  18 - 24 mo &
   &
  \begin{tabular}[c]{@{}l@{}}0.362 (0.042)\end{tabular} &
  - &
  \begin{tabular}[c]{@{}l@{}}0.421 (0.047)\end{tabular} &
  \begin{tabular}[c]{@{}l@{}}0.536 (0.054)\end{tabular} &
  - &
  \textbf{\begin{tabular}[c]{@{}l@{}}0.632 (0.037)\end{tabular}} \\ \cline{2-10} 
 &
  \multirow{4}{*}{\begin{tabular}[c]{@{}l@{}}$EDSS_{mean}$\\ as severity\\ category\end{tabular}} &
  0 - 6 mo &
  \multirow{4}{*}{\begin{tabular}[c]{@{}l@{}}Avg.\\ AU PRC\end{tabular}} &
  \begin{tabular}[c]{@{}l@{}}0.523 (0.015)\end{tabular} &
  - &
  \begin{tabular}[c]{@{}l@{}}0.687 (0.01)\end{tabular} &
  \begin{tabular}[c]{@{}l@{}}0.717 (0.015)\end{tabular} &
  - &
  \textbf{\begin{tabular}[c]{@{}l@{}}0.782 (0.028)\end{tabular}} \\ \cline{3-3} \cline{5-10} 
 &
   &
  6 - 12 mo &
   &
  \begin{tabular}[c]{@{}l@{}}0.47 (0.015)\end{tabular} &
  - &
  \begin{tabular}[c]{@{}l@{}}0.633 (0.018)\end{tabular} &
  \begin{tabular}[c]{@{}l@{}}0.675 (0.016)\end{tabular} &
  - &
  \textbf{\begin{tabular}[c]{@{}l@{}}0.709 (0.044)\end{tabular}} \\ \cline{3-3} \cline{5-10} 
 &
   &
  12 - 18 mo &
   &
  \begin{tabular}[c]{@{}l@{}}0.434 (0.017)\end{tabular} &
  - &
  \begin{tabular}[c]{@{}l@{}}0.606 (0.012)\end{tabular} &
  \begin{tabular}[c]{@{}l@{}}0.649 (0.019)\end{tabular} &
  - &
  \textbf{\begin{tabular}[c]{@{}l@{}}0.674 (0.037)\end{tabular}} \\ \cline{3-3} \cline{5-10} 
 &
   &
  18 - 24 mo &
   &
  \begin{tabular}[c]{@{}l@{}}0.413 (0.017)\end{tabular} &
  - &
  \begin{tabular}[c]{@{}l@{}}0.575 (0.02)\end{tabular} &
  \begin{tabular}[c]{@{}l@{}}0.625 (0.02)\end{tabular} &
  - &
  \textbf{\begin{tabular}[c]{@{}l@{}}0.632 (0.037)\end{tabular}} \\ \hline
\end{tabular}%
}
\end{table*}

\subsubsection{Floodlight}

\noindent \textbf{Performance on disability scores:}
Table \ref{tab:floodlight_full_feature_results} reports model performance on disability scores defined on the Floodlight dataset (see Sec.\ \ref{floodlight_disability_scores}). For Floodlight, we observe that Gradient Boosted Regressor outperforms the other models in all 12 tasks. We believe that this is due to the fact that Floodlight endpoints are zero-inflated (see Appendix \ref{prediction_task_stats}, Table A\ref{tab:label_stats}) and we intend to explore zero-inflated versions of the models in future. The standard deviations show that the obtained results have tight bounds. Note that the inclusion of multiple families of models allows us to explore and find the best model per dataset or task. Performance on the Floodlight dataset remains relatively similar across different time horizons, potentially since these are not as further out as MSOAC.

\noindent \textbf{Performance on smartphone-based MS diagnosis:}
We summarize the results of diagnosing MS in this dataset (see Sec.\ \ref{smartphone_ms_diagnosis}) using smartphone-based tests in Figure \ref{ms_vs_non_ms}. The results demonstrate that overall TCN achieves the best performance across various values of $N$ (number of tests) and shows 71.67\% accuracy at $N=100$ (approximately 1.5 weeks of app usage). Moreover, while for all models the performance improves as $N$ increases, the improvement is highest for TCN (8.89\% increase from $N$=5 to $N$=100) and lowest for Logistic Regression (1.95\% increase from $N$=5 to $N$=100).

\subsubsection{Cross-dataset}

\noindent \textbf{Performance on disability progression labels:}
Next we look at the disability progression labels described in Sec.\ \ref{subsec:disab_prog_labels} for both the MSOAC and Floodlight dataset. For MSOAC we focus only on TCN, since TCN outperforms all other models for the previous endpoints (see Table \ref{tab:msoac_full_feature_results}). We observe that the disability progression tasks have a high class imbalance (92.86--98.27\%), and hence we use both cross-entropy loss and focal loss (popular for imbalanced datasets) \citep{lin2018focal} during training. The latter provides a much better performance for this task, with an average AU PRC improvement of 10.27\%. The mean and standard deviation of AU PRC are reported in Table \ref{disab_prog_TCN}. Note that we also explored focal loss for the classification labels listed in Table \ref{tab:msoac_full_feature_results}, however there was no observable improvement, potentially since these endpoints do not show high imbalance.

While results are promising on MSOAC, the models were not able to obtain significant results using Floodlight. This result might be due to the daily bucketing of measurements, which leads to a label prevalence lower than 0.01\%. Weekly bucketing could be considered in the future.

\begin{table*}[h]
\centering
\caption{Performance (and standard deviation) obtained by machine learning models on a diverse set of prediction tasks for the Floodlight dataset. Best results are reported in bold.}
\label{tab:floodlight_full_feature_results}
\resizebox{0.8\textwidth}{!}{%
\begin{tabular}{|l|l|l|l|l|l|l|}
\hline
\textbf{Dataset} &
  \begin{tabular}[c]{@{}l@{}}\textbf{Prediction}\\ \textbf{tasks}\end{tabular} &
  \begin{tabular}[c]{@{}l@{}}\textbf{Prediction}\\ \textbf{Window}\end{tabular} &
  \textbf{Metric} &
  \begin{tabular}[c]{@{}l@{}}\textbf{Linear}\\ \textbf{Regression}\end{tabular} &
  \begin{tabular}[c]{@{}l@{}}\textbf{Gradient}\\ \textbf{Boosted}\\ \textbf{Regressor}\end{tabular} &
  \begin{tabular}[c]{@{}l@{}}\textbf{TCN}\end{tabular} \\ \hline
\multirow{4}{*}{Floodlight} &
  \multirow{3}{*}{\begin{tabular}[c]{@{}l@{}}Cognitive\\ disability\\ score\end{tabular}} &
  0 - 1 wk &
  \multirow{3}{*}{R-MSE} &
  \begin{tabular}[c]{@{}l@{}}0.275 (0.015)\end{tabular} &
  \textbf{\begin{tabular}[c]{@{}l@{}}0.262 (0.016)\end{tabular}} &
  \begin{tabular}[c]{@{}l@{}}0.313 (0.018)\end{tabular} \\ \cline{3-3} \cline{5-7} 
 &
   &
  1 - 2 wks &
   &
  \begin{tabular}[c]{@{}l@{}}0.285 (0.015)\end{tabular} &
  \textbf{\begin{tabular}[c]{@{}l@{}}0.275 (0.014)\end{tabular}} &
  \begin{tabular}[c]{@{}l@{}}0.337 (0.025)\end{tabular} \\ \cline{3-3} \cline{5-7} 
 &
   &
  2 - 4 wks &
   &
  \begin{tabular}[c]{@{}l@{}}0.286 (0.014)\end{tabular} &
  \textbf{\begin{tabular}[c]{@{}l@{}}0.279 (0.013)\end{tabular}} &
  \begin{tabular}[c]{@{}l@{}}0.353 (0.023)\end{tabular} \\ \cline{2-7} 
 &
  \multirow{3}{*}{\begin{tabular}[c]{@{}l@{}}Dexterity\\ disability\\ score\end{tabular}} &
  0 - 1 wk &
  \multirow{3}{*}{R-MSE} &
  \begin{tabular}[c]{@{}l@{}}0.152 (0.011)\end{tabular} &
  \textbf{\begin{tabular}[c]{@{}l@{}}0.146 (0.012)\end{tabular}} &
  \begin{tabular}[c]{@{}l@{}}0.162 (0.018)\end{tabular} \\ \cline{3-3} \cline{5-7} 
 &
   &
  1 - 2 wks &
   &
  \begin{tabular}[c]{@{}l@{}}0.153 (0.012)\end{tabular} &
  \textbf{\begin{tabular}[c]{@{}l@{}}0.148 (0.012)\end{tabular}} &
  \begin{tabular}[c]{@{}l@{}}0.171 (0.022)\end{tabular} \\ \cline{3-3} \cline{5-7} 
 &
   &
  2 - 4 wks &
   &
  \begin{tabular}[c]{@{}l@{}}0.152 (0.011)\end{tabular} &
  \textbf{\begin{tabular}[c]{@{}l@{}}0.149 (0.011)\end{tabular}} &
  \begin{tabular}[c]{@{}l@{}}0.178 (0.019)\end{tabular} \\ \cline{2-7} 
 &
  \multirow{3}{*}{\begin{tabular}[c]{@{}l@{}}Mobility\\ disability\\ score\end{tabular}} &
  0 - 1 wk &
  \multirow{3}{*}{R-MSE} &
  \begin{tabular}[c]{@{}l@{}}0.244 (0.017)\end{tabular} &
  \textbf{\begin{tabular}[c]{@{}l@{}}0.226 (0.018)\end{tabular}} &
  \begin{tabular}[c]{@{}l@{}}0.283 (0.021)\end{tabular} \\ \cline{3-3} \cline{5-7} 
 &
   &
  1 - 2 wks &
   &
  \begin{tabular}[c]{@{}l@{}}0.256 (0.018)\end{tabular} &
  \textbf{\begin{tabular}[c]{@{}l@{}}0.244 (0.021)\end{tabular}} &
  \begin{tabular}[c]{@{}l@{}}0.313 (0.026)\end{tabular} \\ \cline{3-3} \cline{5-7} 
 &
   &
  2 - 4 wks &
   &
  \begin{tabular}[c]{@{}l@{}}0.26 (0.017)\end{tabular} &
  \textbf{\begin{tabular}[c]{@{}l@{}}0.249 (0.019)\end{tabular}} &
  \begin{tabular}[c]{@{}l@{}}0.328 (0.027)\end{tabular} \\ \cline{2-7} 
 &
  \multirow{3}{*}{\begin{tabular}[c]{@{}l@{}}Overall\\ disability\\ score\end{tabular}} &
  0 - 1 wk &
  \multirow{3}{*}{R-MSE} &
  \begin{tabular}[c]{@{}l@{}}0.192 (0.012)\end{tabular} &
  \textbf{\begin{tabular}[c]{@{}l@{}}0.18 (0.013)\end{tabular}} &
  \begin{tabular}[c]{@{}l@{}}0.225 (0.019)\end{tabular} \\ \cline{3-3} \cline{5-7} 
 &
   &
  1 - 2 wks &
   &
  \begin{tabular}[c]{@{}l@{}}0.206 (0.012)\end{tabular} &
  \textbf{\begin{tabular}[c]{@{}l@{}}0.197 (0.013)\end{tabular}} &
  \begin{tabular}[c]{@{}l@{}}0.246 (0.017)\end{tabular} \\ \cline{3-3} \cline{5-7} 
 &
   &
  2 - 4 wks &
   &
  \begin{tabular}[c]{@{}l@{}}0.209 (0.011)\end{tabular} &
  \textbf{\begin{tabular}[c]{@{}l@{}}0.209 (0.011)\end{tabular}} &
  \begin{tabular}[c]{@{}l@{}}0.265 (0.019)\end{tabular} \\ \hline
\end{tabular}%
}
\end{table*}

\begin{table}[h]
\centering
\caption{Performance (and standard deviation) of TCN on functional test specific disability progression labels defined on the MSOAC dataset.}
\label{disab_prog_TCN}
\resizebox{0.45\textwidth}{!}{%
\begin{tabular}{|c|c|cc|}
\hline
\multirow{2}{*}{\begin{tabular}[c]{@{}c@{}}Functional\\ test\end{tabular}} &
  \multirow{2}{*}{\begin{tabular}[c]{@{}c@{}}Prediction\\ horizon\end{tabular}} &
  \multicolumn{2}{c|}{Loss type} \\ \cline{3-4} 
                       &         & \multicolumn{1}{c|}{\begin{tabular}[c]{@{}c@{}}Cross-entropy\\ loss\end{tabular}} & Focal loss    \\ \hline
\multirow{2}{*}{NHPT}  & 0-6 mo  & \multicolumn{1}{c|}{0.406 (0.030)}                                                & \textbf{0.583 (0.142)} \\ \cline{2-4} 
                       & 6-12 mo & \multicolumn{1}{c|}{0.396 (0.021)}                                                & \textbf{0.534 (0.171)} \\ \hline
\multirow{2}{*}{PASAT} & 0-6 mo  & \multicolumn{1}{c|}{0.476 (0.016)}                                                & \textbf{0.533 (0.172)} \\ \cline{2-4} 
                       & 6-12 mo & \multicolumn{1}{c|}{0.482 (0.012)}                                                & \textbf{0.567 (0.161)} \\ \hline
\multirow{2}{*}{SDMT}  & 0-6 mo  & \multicolumn{1}{c|}{0.471 (0.058)}                                                & \textbf{0.522 (0.112)}  \\ \cline{2-4} 
                       & 6-12 mo & \multicolumn{1}{c|}{0.534 (0.084)}                                                & \textbf{0.535 (0.158)} \\ \hline
\multirow{2}{*}{T25FW} & 0-6 mo  & \multicolumn{1}{c|}{0.422 (0.017)}                                                & \textbf{0.558 (0.147)} \\ \cline{2-4} 
                       & 6-12 mo & \multicolumn{1}{c|}{0.423 (0.015)}                                                & \textbf{0.600 (0.053)} \\ \hline
\end{tabular}%
}
\end{table}

\subsection{Feature ablation}

To assess the signal related to disease progression in each feature group, we perform different feature ablation experiments. For both MSOAC and Floodlight we choose the following groups: demographics, questionnaires, functional tests, and all POMs (questionnaires + functional tests).

The results are presented in Table \ref{msoac_feature_ablation_main}. Due to space constraints, we choose a single prediction horizon per dataset (6-12 months for MSOAC and 1-2 weeks for Floodlight), for all the tasks considered in Tables \ref{tab:msoac_full_feature_results} and \ref{tab:floodlight_full_feature_results}. While in Table \ref{msoac_feature_ablation_main} we show the results for only the top performing model per dataset (TCN for MSOAC and Gradient Booster Regressor for Floodlight), results for all models are reported in Table A\ref{feature_ablation_appendix} (Appendix). The trends obtained for the top performing models are consistent across other models as well.

For MSOAC, the results depict that the feature groups are of following importance: demographics $<$ questionnaires $<$ functional tests $<$ all POMs $<$ full feature set. In line with expectations, the full feature set containing both POMs and demographic features produces the best performance. Demographics (static features) impact model performance the least, and are outperformed by functional tests and questionnaires (dynamic/temporal features). Between functional tests and questionnaires, we observe that the former outperforms the latter. We believe the reason is evident from Figure \ref{feature_sparsity_msoac} which shows that the questionnaires are orders of magnitude sparser than functional tests thereby leading to less signal for the ML models. 

For Floodlight, the order of importance of feature groups is as follows: questionnaires $<$ demographics $<$ functional tests $<$ all POMs $<$ full feature set. Compared to MSOAC, the importance of questionnaires and demographics have flipped. This result is expected for Floodlight since it contains only one questionnaire feature (Mood Response) unlike MSOAC consisting of multiple questionnaire features. 

\begin{table*}[h]
\centering
\caption{Summary of feature ablation studies on MSOAC and Floodlight for 6-12 months and 1-2 weeks horizon respectively for the best performing models from Table \ref{tab:msoac_full_feature_results} and \ref{tab:floodlight_full_feature_results} respectively. For MSOAC, will $EDSS_{mean}$ reports R-MSE (lower better), all other labels report AU PRC (higher better). For Floodlight, R-MSE is reported for all labels.}
\label{msoac_feature_ablation_main}
\resizebox{\textwidth}{!}{%
\begin{tabular}{|c|l|cclcc|}
\hline
\multirow{2}{*}{\textbf{Dataset}} &
  \multicolumn{1}{c|}{\multirow{2}{*}{\textbf{\begin{tabular}[c]{@{}c@{}}Prediction\\ tasks\end{tabular}}}} &
  \multicolumn{5}{c|}{\textbf{Feature Groups}} \\ \cline{3-7} 
 &
  \multicolumn{1}{c|}{} &
  \multicolumn{1}{c|}{\textit{Demographics}} &
  \multicolumn{1}{c|}{\textit{\begin{tabular}[c]{@{}c@{}}Functional\\ Tests\end{tabular}}} &
  \multicolumn{1}{l|}{\textit{Questionnaires}} &
  \multicolumn{1}{c|}{\textit{\begin{tabular}[c]{@{}c@{}}Performance\\ Outcome\\ Measures\end{tabular}}} &
  \textit{Full feature set} \\ \hline
\multirow{4}{*}{MSOAC} &
  $EDSS_{mean}$ &
  \multicolumn{1}{c|}{1.957 (0.051)} &
  \multicolumn{1}{c|}{1.777 (0.091)} &
  \multicolumn{1}{l|}{1.860 (0.049)} &
  \multicolumn{1}{c|}{1.676 (0.077)} &
  \textbf{1.650 (0.067)} \\ \cline{2-7} 
 &
  \shortstack{\\$EDSS_{mean} > 3$ \\ (Moderate disability)} &
  \multicolumn{1}{c|}{0.678 (0.019)} &
  \multicolumn{1}{c|}{0.766 (0.025)} &
  \multicolumn{1}{l|}{0.789 (0.020)} &
  \multicolumn{1}{c|}{0.816 (0.037)} &
  \textbf{0.820 (0.027)} \\ \cline{2-7} 
 &
  \shortstack{\\$EDSS_{mean} > 5$ \\ (Severe disability)} &
  \multicolumn{1}{c|}{0.456 (0.028)} &
  \multicolumn{1}{c|}{0.665 (0.036)} &
  \multicolumn{1}{l|}{0.608 (0.037)} &
  \multicolumn{1}{c|}{0.691 (0.034)} &
  \textbf{0.722 (0.039)} \\ \cline{2-7} 
 &
  \shortstack{\\$EDSS_{mean}$ as severity \\ category} &
  \multicolumn{1}{c|}{0.520 (0.011)} &
  \multicolumn{1}{c|}{0.672 (0.086)} &
  \multicolumn{1}{l|}{0.659 (0.016)} &
  \multicolumn{1}{c|}{0.686 (0.042)} &
  \textbf{0.709 (0.044)} \\ \hline
\multirow{4}{*}{Floodlight} &
  \shortstack{Cognitive\\ disability score} &
  \multicolumn{1}{c|}{0.306 (0.017)} &
  \multicolumn{1}{c|}{0.286 (0.012)} &
  \multicolumn{1}{l|}{0.416 (0.037)} &
  \multicolumn{1}{c|}{0.283 (0.014)} &
  \textbf{0.275 (0.014)} \\ \cline{2-7} 
 &
  \shortstack{Dexterity\\ disability score} &
  \multicolumn{1}{c|}{0.159 (0.014)} &
  \multicolumn{1}{c|}{0.153 (0.012)} &
  \multicolumn{1}{l|}{0.198 (0.021)} &
  \multicolumn{1}{c|}{0.152 (0.012)} &
  \textbf{0.148 (0.012)} \\ \cline{2-7} 
 &
  \shortstack{Mobility\\ disability score} &
  \multicolumn{1}{l|}{0.278 (0.018)} &
  \multicolumn{1}{l|}{0.249 (0.017)} &
  \multicolumn{1}{l|}{0.381 (0.031)} &
  \multicolumn{1}{l|}{0.248 (0.019)} &
  \multicolumn{1}{l|}{\textbf{0.244 (0.021)}} \\ \cline{2-7} 
 &
  \shortstack{Overall\\ disability score} &
  \multicolumn{1}{l|}{0.220 (0.012)} &
  \multicolumn{1}{l|}{0.206 (0.012)} &
  \multicolumn{1}{l|}{0.308 (0.034)} &
  \multicolumn{1}{l|}{0.205 (0.013)} &
  \multicolumn{1}{l|}{\textbf{0.197 (0.013)}} \\ \hline
\end{tabular}%
}
\end{table*}

\subsection{Subgroup results}
\label{sec:subgroup_perf}
Apart from performance on the entire dataset, subgroup analysis enables researchers and clinicians to understand where models fall short.

For both datasets we look at the sex and age-bucketed subgroups. Given that MS is a disease that tends to affect more women than men, we assess potential discrepancies in model performance between males and females. Stratification on age is also a relevant evaluation, as MS is a long-term condition and younger patients typically have a less severe form of the disease. As the disease progresses, it can evolve from relapse-remitting to a progressive state \citep{10.3389/fneur.2021.608491}, which is often accompanied by an increase in symptom severity. Our age buckets were chosen based on expert input on what would be most clinically useful.

Table A\ref{msoac_subgroup_results} (Appendix) contains the results on 3 predictions tasks for MSOAC, for the 6-12 month horizon on all models. We only report results where the subgroup was known, as this information is not present for all patients.

We see that for both males and females the models tend to have a similar, and sometimes identical performance (AU PRC), across all folds.  Age on the other hand sees a discrepancy when it comes to the various subgroups, with people aged under 30 seeing the biggest decrease in AU PRC. For people aged 50-70 we see that performance is either on par (EDSS \textgreater 3, EDSS \textgreater 5) or slightly lower (EDSS as severity category) to that on the full dataset. This is surprising, as the overall composition of MSOAC is relapse-remitting patients and we would expect people in this age group to have a more stable form of the disease. We note that evaluation for patients aged 70 or older is not informative given the low number of cases (n=11) in this group. A distribution of label values based on age can be found in Appendix \ref{prediction_task_stats}.

\section{Discussion and future work}
\label{sec:discussion}
In this paper, we show for the first time that it is possible to predict disease progression in MS using POMs, demographic information, and machine learning for both a clinical trial and smartphone-based dataset. Early prediction of disability in both settings has the potential to support MS subjects and healthcare professionals, since the best course of action is early diagnosis and symptom treatment. This in turn can lead to a slower disease progression and a better quality of life for a longer period of time \citep{Cerqueira844}. Smartphone-based monitoring can also enable early diagnosis of MS.

Temporal patterns of POMs seem to play an important role in the prediction of longer term disability, as shown by the better performance of TCN in all tasks in the MSOAC dataset. Similarly, results on Floodlight display that short-term patterns (a couple of weeks) also carry predictive power. These results suggest that the continuous evaluation of POMs is a promising avenue for the monitoring and early detection of disease progression of MS patients. Both long- and short-term predictions are potentially clinically actionable since while the former may lead to individually-tailored disease-modifying therapies \citep{pmid30766181}, the latter may prompt a clinician to prescribe medications to control symptoms \citep{nmss_symptom}. 


While the full feature set leads to the highest model performance (according to AU PRC) when compared to feature ablations, we note that POMs without demographics perform on par. This result, although preliminary, questions the recording of demographic data for predictive purposes. Future work could further investigate whether demographic data is indeed beneficial in terms of model performance and patient outcome, considering the balance between data need and privacy. Moreover, the relatively higher sparsity of questionnaires compared to functional tests and its eventual impact in model performance points toward the need of collecting more user-friendly questionnaires, more reliably, and over a longer horizon.

The limitations of this study include (i) the disability scores defined for Floodlight, while inspired from EDSS, are experimental and would require further clinical validation to ensure soundness and clinical relevance, and (ii) the disability progression labels (Sec.\ \ref{subsec:disab_prog_labels}) did not lead to a well-defined machine learning problem for the Floodlight dataset due to a high-frequency of short-term recordings.

The plans for future work for this study are multi-fold. First, we shall ingest more relevant features (e.g. medication, medical history), in addition to the current POMs and demographic data. Second, we intend to explore different time-bucketing techniques for Floodlight to tackle the imbalance of disability progression labels and more closely relate to clinical actionability. Third, we intend to handle the irregularity of the features by continuous time modelling instead of missing-value imputation \citep{kidger2020neural}. Fourth, we plan to further evaluate the robustness of our models, especially across longer time horizons. Fifth, we intend to create a large-scale multi-site smartphone-based MS dataset for further evaluation and potential deployment of the developed models. As discussed in Sec. \ref{sec:subgroup_perf}, we lack data in specific patient subgroups. Targeted data acquisition in e.g.\ patients over 70 could be considered for model evaluation, and potentially for model training.

\section*{Institutional Review Board (IRB)}
Our research does not require IRB approval, since the datasets are publicly available.

\acks{We would like to thank the Multiple Sclerosis Outcome Assessments Consortium and Genentech for making the two multiple sclerosis datasets publicly available for research.}

\bibliography{refs}

\appendix

\section{Subject representation and MS dataset conversion}
\label{subject_proto_details}
As described in Sec.\ \ref{subsec:data_processing}, we convert the MS datasets into a common representation called Subject described in Table \ref{subject_representation}. The Subject representation allows us to map a diverse set of datasets into a common format consisting a pre-defined set of fields. This allows not only for an easier downstream processing of multiple datasets, but potentially joining multiple datasets into one.

The fields in Subject are chosen in a way that it stores all information relevant to MS datasets (both clinical and at-home), and to generalize to other healthcare datasets as well. The representation can also be expanded to include unique dataset-specific intricacies.

\begin{table*}[!h]
\centering
\caption{Description of the common Subject representation. }
\label{subject_representation}
\resizebox{\textwidth}{!}{%
\begin{tabular}{|l|l|l|l|}
\hline
\textbf{Field}                 & \textbf{Type}                             & \textbf{Format} & \textbf{Description}                                            \\ \hline
\multicolumn{4}{|l|}{\textbf{Subject}: Defines all the data provided for a single subject.}                                                                    \\ \hline
subject\_id                    & string                                    & Optional        & Unique ID for each subject.                                     \\ \hline
subject\_characteristics       & SubjectCharacteristics                    & Optional        & Defines the subject's characteristics.                          \\ \hline
medical\_history               & MedicalHistory                            & Optional        & Defines the subject's medical history.                          \\ \hline
episodes                       & Episode                                   & Repeated        & A sequence of encounters corresponding to a single subject.     \\ \hline
\multicolumn{4}{|l|}{\textbf{SubjectCharacteristics}: Defines all the data provided for a single subject.}                                                              \\ \hline
sex                            & Sex.Enum                                  & Optional        & The subject's sex.                                              \\ \hline
\multicolumn{4}{|l|}{\begin{tabular}[c]{@{}l@{}}\textbf{MedicalHistory}: Defines a sequence of outpatient events recorded before the current system started recording events.\\ These typically are recorded with the same timestamp even though they took place over longer periods of time.\end{tabular}} \\ \hline
clinical\_events               & ClinicalEvent                             & Repeated        & A clinical event.                                               \\ \hline
\multicolumn{4}{|l|}{\textbf{Episode}: An abstraction of an event sequence.}                                                                                            \\ \hline
clinical\_event                & ClinicalEvent                             & Repeated        & A sequence of clinical events.                                  \\ \hline
changing\_characteristics      & ChangingSubjectCharacteristics            & optional        & Defines the subject's changing characteristics.                 \\ \hline
clinical\_trial &
  ClinicalTrial &
  optional &
  \begin{tabular}[c]{@{}l@{}}Details about the clinical trial the subject was part of\\ (if applicable).\end{tabular} \\ \hline
\multicolumn{4}{|l|}{\textbf{ChangingSubjectCharacteristics}: Defines the subject's changing characteristics.}                                                          \\ \hline
age                            & float                                     & optional        & The subject's age.                                              \\ \hline
gender                         & Gender.Enum                               & optional        & The subject's gender                                            \\ \hline
race                           & string                                    & optional        & The subject's race.                                             \\ \hline
weight                         & float                                     & optional        & The subject's weight.                                           \\ \hline
height                         & float                                     & optional        & The subject's height.                                           \\ \hline
country                        & string                                    & optional        & The subject's country.                                          \\ \hline
\multicolumn{4}{|l|}{\textbf{Sex.Enum}: Defines the subject's sex information.}                                                                                         \\ \hline
FEMALE                         & 0                                         & -               & Female sex.                                                     \\ \hline
MALE                           & 1                                         & -               & Male sex.                                                       \\ \hline
\multicolumn{4}{|l|}{\textbf{Gender.Enum}: Defines the subject's gender information.}                                                                                   \\ \hline
UNKNOWN                        & 0                                         & -               & Unknown gender.                                                 \\ \hline
FEMALE                         & 1                                         & -               & Female gender.                                                  \\ \hline
MALE                           & 2                                         & -               & Male gender.                                                    \\ \hline
OTHER                          & 3                                         & -               & Other gender.                                                   \\ \hline
\multicolumn{4}{|l|}{\textbf{ClinicalEvent}: Defines a single event or a set of coinciding individual events that happen at the same time.} \\ \hline
timestamp &
  int64 &
  optional &
  \begin{tabular}[c]{@{}l@{}}Timestamp of event. While for MSOAC, this corresponds to\\ the day of the clinic visit, for Floodlight, this is the\\ timestamp recorded by the smartphone.\end{tabular} \\ \hline
resources                      & Resource                                  & repeated        & A list of specific clinical entries recorded at this timestamp. \\ \hline
classification\_labels         & map\textless{}string, int64\textgreater{} & required        & Classification labels for prediction.                           \\ \hline
regression\_labels             & map\textless{}string, float\textgreater{} & required        & Regression labels for prediction.                               \\ \hline
\multicolumn{4}{|l|}{Resource: Describes the various types of resources that can be contained within ClinicalEvent.}                                           \\ \hline
functional\_test               & FunctionalTest                            & optional        & Functional assessment test data.                                \\ \hline
questionnaire                  & Questionnaire                             & optional        & Questionnaires filled by the subjects.                          \\ \hline
generic\_resource              & GenericResource                           & optional        & Generic resource to store dataset-specific intricacies.         \\ \hline
\multicolumn{4}{|l|}{\textbf{FunctionalTest}: Functional assessment test data recorded from the subject.}                                                               \\ \hline
name                           & string                                    & optional        & Name of performance outcome measure.                            \\ \hline
category                       & string                                    & optional        & Category of performance outcome measure.                        \\ \hline
response                       & NumericalResponse                         & optional        & Numerical response recorded from subject.                       \\ \hline
\multicolumn{4}{|l|}{\textbf{Questionnaire}: Questionnaire data recorded from the subject.}                                                                             \\ \hline
name                           & string                                    & optional        & Name of performance outcome measure.                            \\ \hline
category                       & string                                    & optional        & Category of performance outcome measure.                        \\ \hline
response                       & QuestionnaireResponse                     & optional        & Questionnaire response recorded from subject.                   \\ \hline
\multicolumn{4}{|l|}{\textbf{NumericalResponse}: Numeric response converted to standardized unit.}                                                                      \\ \hline
numerical\_response\_std\_unit & float                                     & optional        & Numeric response converted to standardized unit.                \\ \hline
std\_unit &
  string &
  optional &
  \begin{tabular}[c]{@{}l@{}}Standardized unit. This unit was used for\\ homogenizing the data.\end{tabular} \\ \hline
\multicolumn{4}{|l|}{\textbf{QuestionnaireResponse}: Responses recorded from questionnaires.}                                                                           \\ \hline
text\_response\_orig           & string                                    & optional        & Response in original text.                                      \\ \hline
text\_response\_std            & string                                    & optional        & Response in standardized text.                                  \\ \hline
numeric\_response\_std         & float                                     & optional        & Standardized numeric response.                                  \\ \hline
categorical\_response\_std &
  string &
  optional &
  \begin{tabular}[c]{@{}l@{}}Standardized categorical response.An example entry\\ is EDSS.\end{tabular} \\ \hline
\end{tabular}%
}
\end{table*}

\begin{table*}[!h]
\centering
\caption{Description of the common Prediction representation. }
\label{prediction_representation}
\resizebox{\textwidth}{!}{%
\begin{tabular}{|l|l|l|}
\hline
\textbf{Field}  & \textbf{Type}   & \textbf{Description} \\ \hline
\multicolumn{3}{|l|}{\textbf{Prediction}: Defines the model output information.} \\ \hline
subject\_id  & string  & Unique ID for each subject. \\ \hline
timestamp & int64 & \begin{tabular}[c]{@{}l@{}}Timestamp of event. While for MSOAC, this corresponds \\ to the day of the clinic visit, for Floodlight, this is the\\ timestamp recorded by the smartphone.\end{tabular}. \\ \hline
label\_targets & map\textless{}string, float\textgreater{}   & \begin{tabular}[c]{@{}l@{}}A mapping from label name to the target value for the \\ particular timestamp this is recorded for. \end{tabular} \\ \hline
label\_predictions  & map\textless{}string, list\textless{}float\textgreater{}\textgreater{} & \begin{tabular}[c]{@{}l@{}}A mapping from label name to the predicted values for \\ the particular timestamp this is recorded for. Multiple \\ values are recorded to account for multi-class predictions.\end{tabular} \\ \hline
subgroup\_attributes & map\textless{}string, oneof\textless{}string, int, float\textgreater{}\textgreater{} & \begin{tabular}[c]{@{}l@{}}A mapping from subgroup name (such as Sex, or Age) to \\ the exact value (e.g. Female, or 56). \end{tabular} \\ \hline
\end{tabular}%
}
\end{table*}

The Prediction format is described in Table \ref{prediction_representation}, and while it's simplistic in its setup, it enables the downstream metrics pipeline development and cross-model comparisons. While in its current form it mainly focuses on time series (through the use of the timestamp field), we believe that it can be adapted to other types as well by simply ignoring the time value. We chose to store a series of label targets and predictions at once, to ease at-scale metrics computations. Thus, our pipeline computes a variety of metrics for all tasks at once.

\section{Feature statistics}
\label{appendix:feature_stats}

\begin{figure}[!ht]
\begin{center}
\centerline{\includegraphics[width=1\columnwidth]{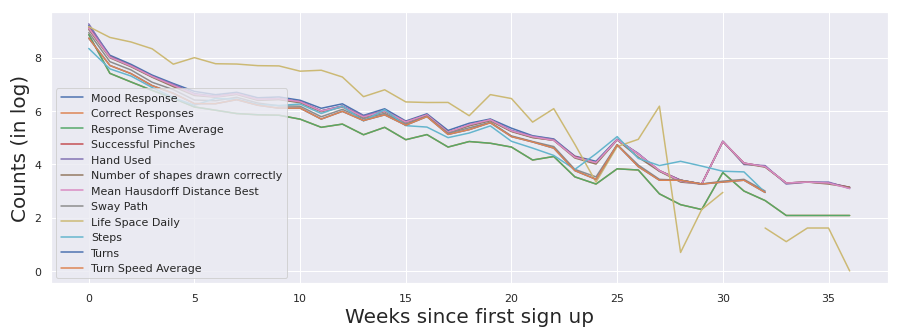}}
\caption{Feature sparsity in Floodlight}
\label{feature_sparsity_floodlight}
\end{center}
\end{figure}

Figure \ref{feature_sparsity_floodlight} shows a summary of the Floodlight functional tests and questionnaires, and how often they appear at a population level, when each patient sequence is bucketed into 1-week intervals. The start of the sequence is when they joined the app.

We observe a similar trend to that on MSOAC, with time playing an important factor in feature sparsity. As mentioned in the main text, both functional tests and questionnaires follow a similar pattern due to the simplicity of questionnaires (only one mood related question). This is consistent with the patient drop-off that takes place in Floodlight, which in turn is consistent with other mobile-based studies.

We believe a focus on preventing attrition is needed for an increase in mobile dataset quality, as these types of datasets have the power to harness diverse information at-scale.

\section{Hyperparameter search space}
\label{appendix:hyperparameters}
The hyperparameter search space for each model can be found in Table \ref{tab:hyperparam_search_space}. Future work will look into using parameter auto-tuning tools such as Vizier \citep{Golovin2017}, in order to expand the search space and identify the optimal set-up for our suite of models.

\begin{table*}[]
\centering
\caption{Hyperparameter search for models considered in this study.}
\label{tab:hyperparam_search_space}
\resizebox{0.6\textwidth}{!}{%
\begin{tabular}{|l|l|}
\hline
\textbf{Model    }           & \textbf{Hyperparameter search space }         \\ \hline
Logistic Regression & C = {[}1e-2, 1e-1, 1., 1e+1, 1e+2{]} \\ \hline
Linear Regression   & -                                    \\ \hline
MLP                         & \begin{tabular}[c]{@{}l@{}}network\_size = {[}(16, 16), (16, 16, 16), (32,  32){]}\\ learning\_rate = {[}0.001, 0.01{]}\end{tabular}  \\ \hline
Gradient Boosted Classifier & \begin{tabular}[c]{@{}l@{}}n\_estimators = {[}100, 150{]}\\ learning\_rate = {[}0.001, 0.01{]}\\ max\_depth = {[}3, 5{]}\end{tabular} \\ \hline
Gradient Boosted Regressor  & \begin{tabular}[c]{@{}l@{}}n\_estimators = {[}100, 150{]}\\ learning\_rate = {[}0.001, 0.01{]}\\ max\_depth = {[}3, 5{]}\end{tabular} \\ \hline
TemporalConvNet             & \begin{tabular}[c]{@{}l@{}}num\_filters = {[}16, 32, 64{]}\\ kernel\_size = {[}3, 5{]}\\ learning\_rate = {[}0.001, 0.05, 0.01{]}\\dropout = {[}0.0, 0.5{]}\end{tabular}        \\ \hline
\end{tabular}%
}
\end{table*}

\section{Disability progression labels}
\label{appendix:disability_progression}

\begin{figure}[!ht]
\begin{center}
\centerline{\includegraphics[width=1\columnwidth]{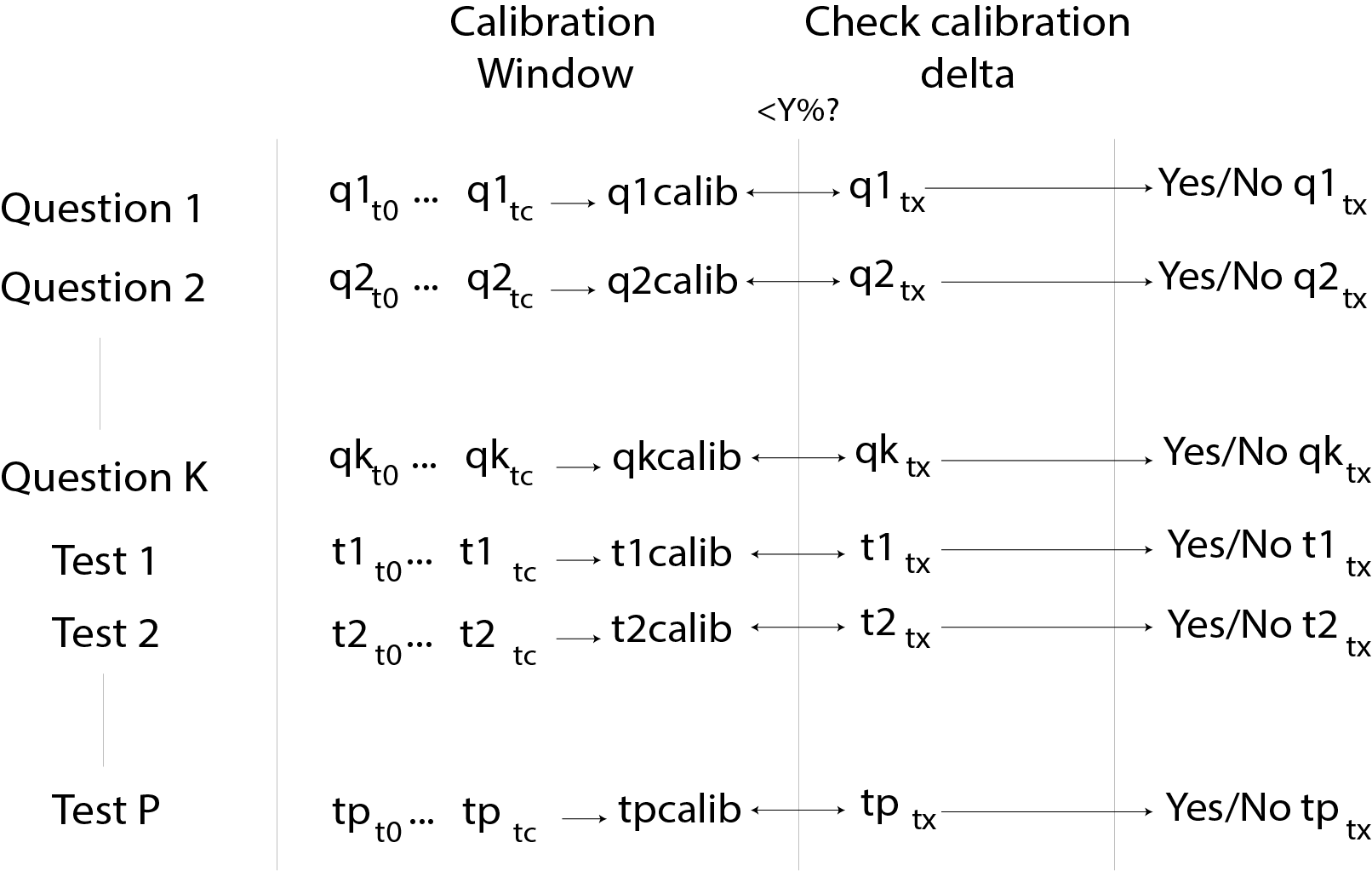}}
\caption{Disability progression label definition}
\label{disability_progression_labels}
\end{center}
\end{figure}

Figure \ref{disability_progression_labels} illustrates the computation of the baseline values and how they are used to create the final Worsening/Unchanged/Improved outcome, for each functional test and questionnaire. In the first \textit{c} timesteps we compute a baseline value for each feature. For each following timestep we perform two actions:
\begin{itemize}
    \item We compare the value at the new timestep with the baseline value we have for that feature. If the difference in value is greater than a threshold (in our case 20\% increase or decrease), we set a label of Worsening/Improving. Otherwise the label value gets set to Unchanged.
    \item We update the baseline value for each feature based on this new information.
\end{itemize}

While this task could have been posed as "are any of the tests/questionnaires deviating from the baseline?", the per-test prediction was considered to be more clinically actionable as the actual test is more informative than "something is wrong".

\section{Statistics of prediction tasks}
\label{prediction_task_stats}

\begin{figure*}[]
  \includegraphics[width=\linewidth]{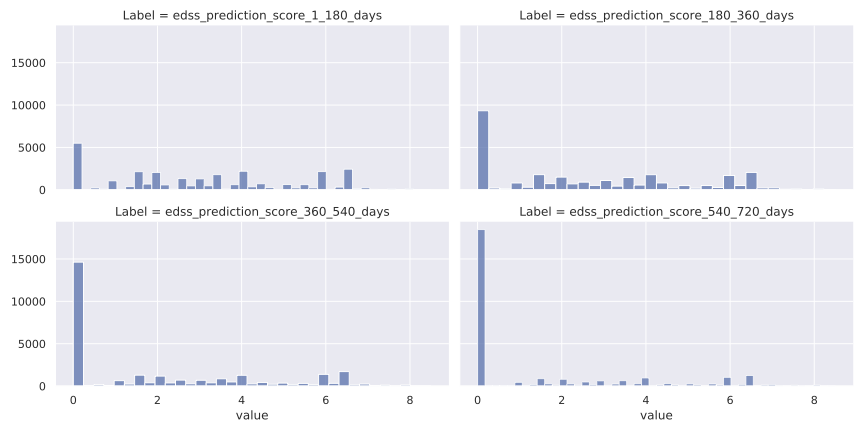}
  \caption{Histograms of regression labels derived from EDSS scores recorded in the MSOAC dataset.}
  \label{msoac_regression_labels}
\end{figure*}

\begin{figure*}[]
  \includegraphics[width=\linewidth]{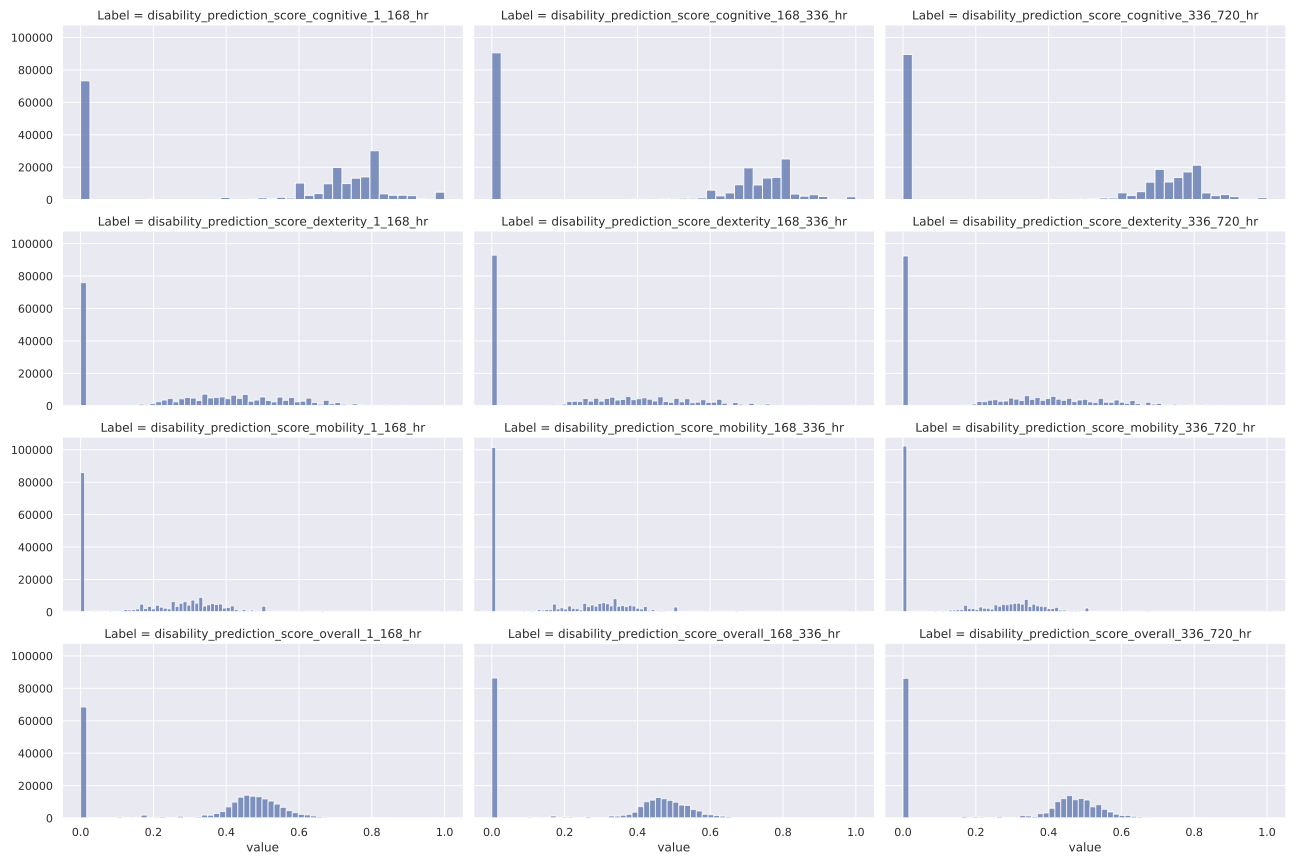}
  \caption{Histograms of regression labels derived from the Floodlight dataset.}
  \label{floodlight_regression_labels}
\end{figure*}

Table \ref{tab:label_stats} shows the class distribution for both binary and multi-class classification tasks, while Figures \ref{msoac_regression_labels} and \ref{floodlight_regression_labels} present a histogram of the label values for the regression tasks.

We note a higher class imbalance for $EDSS_{mean} > 5$, even for the shorter horizon of 0-6 months, while $EDSS_{mean} > 3$ starts off in a more balanced state for the same timeframe. $EDSS_{mean}$ as severity category shows a similar trend to $EDSS_{mean} > 3$, with the shorter time horizon being balanced and the longer time horizon seeing "No disability" as the most prevalent label. We believe this is due to the fact that we do not have information so far into the future, so the default values of "No disability" get used instead. Future work will look into handling this lack of future information. For Floodlight we note that the labels are zero-inflated.

\begin{figure*}[]
  \includegraphics[width=\linewidth]{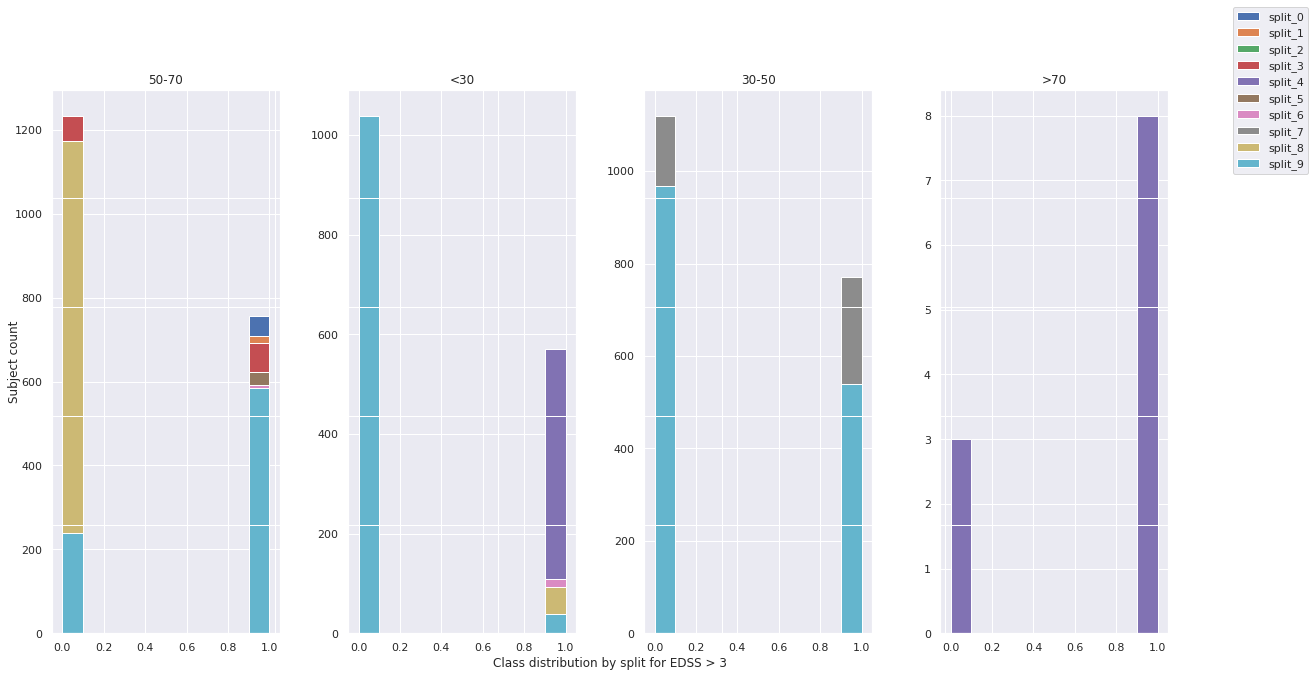}
  \caption{Class distribution by split for EDSS \textgreater 3.}
  \label{age_edss_3}
\end{figure*}

\begin{figure*}[]
  \includegraphics[width=\linewidth]{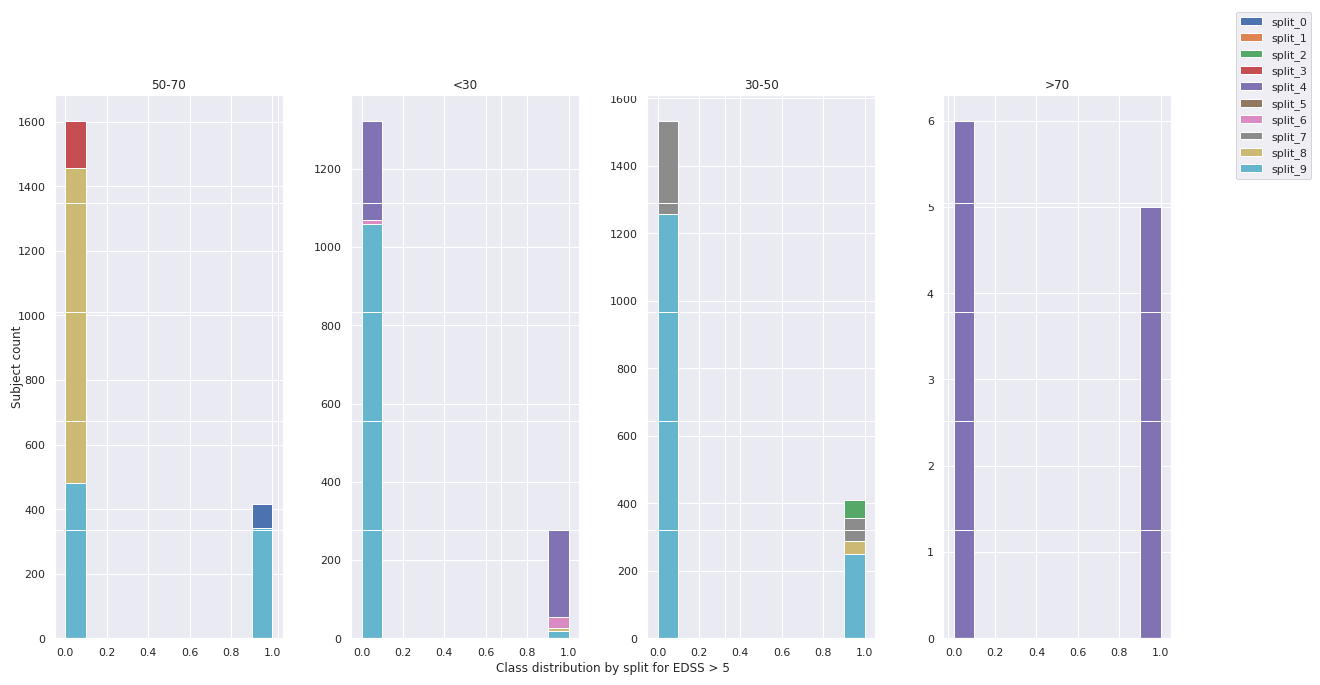}
  \caption{Class distribution by split for EDSS \textgreater 5.}
  \label{age_edss_5}
\end{figure*}

\begin{figure*}[]
  \includegraphics[width=\linewidth]{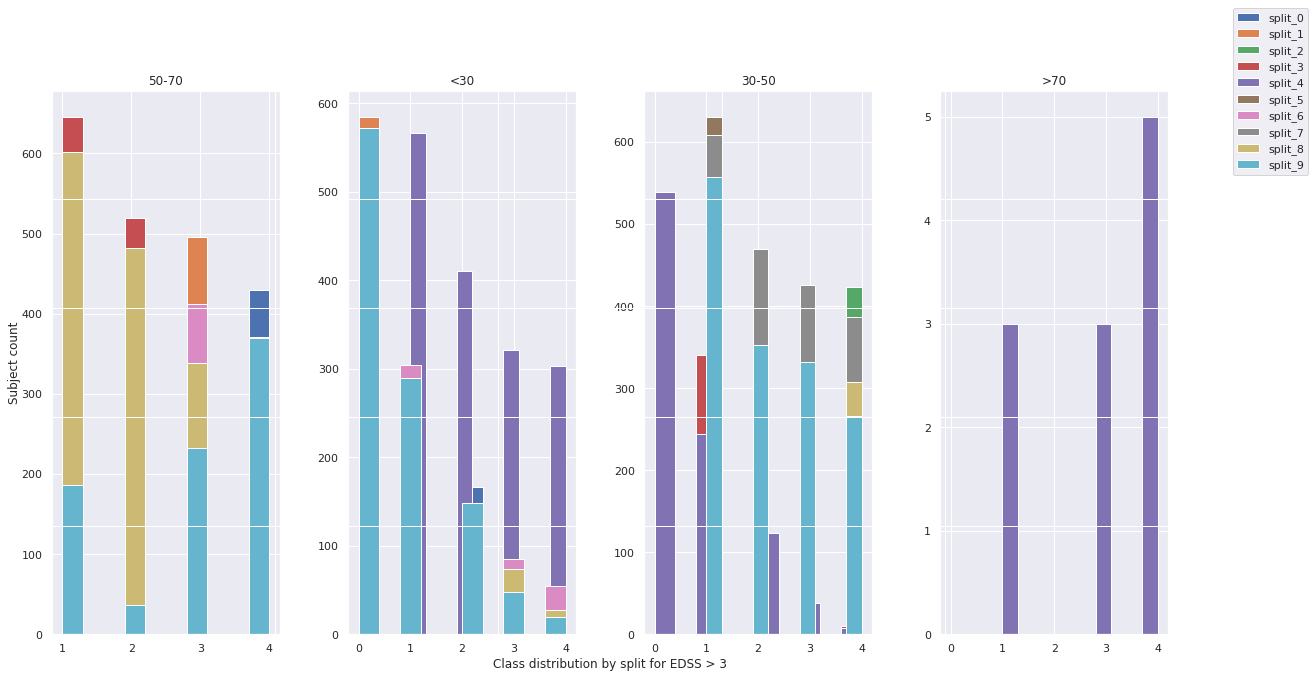}
  \caption{Class distribution by split for EDSS as severity category.}
  \label{age_edss_severity_category}
\end{figure*}

For the classification tasks in MSOAC, we can see the class distribution for the age subgroups, by each cross-validation split, in Figures \ref{age_edss_3}, \ref{age_edss_5} and \ref{age_edss_severity_category}. Note that splits tend to have very different distributions of age groups, which explains the higher standard deviation for the less prominent groups (age \textless{30}).

\begin{table*}[]
\centering
\caption{Label distributions for classification tasks}
\label{tab:label_stats}
{%
    \begin{tabular}{|l|l|l|l|l|}
    \hline
      Dataset &
      \begin{tabular}[c]{@{}l@{}}Prediction\\ tasks\end{tabular} &
      \begin{tabular}[c]{@{}l@{}}Prediction\\ Window\end{tabular} &
      Class &
      Percentage \\ \hline
  \multirow{4}{*}{MSOAC}
  &
    \multirow{4}{*}{\begin{tabular}[c]{@{}l@{}}$EDSS_{mean}$\\ $ > 3$ (Moderate\\ disability)\end{tabular}}
    &
      \multirow{2}{*}{0 - 6 mo}
      &
        False & 53.49 \\
      & & &
        True & 46.51 \\ \cline{3-3} \cline{4-4} \cline{5-5}
    & &
      \multirow{2}{*}{6 - 12 mo}
      &
        False & 60.02 \\
      & & &
        True & 39.98 \\ \cline{3-3} \cline{4-4} \cline{5-5}
    & &
      \multirow{2}{*}{12 - 18 mo}
      &
        False & 69.47 \\
      & & &
        True & 30.53 \\ \cline{3-3} \cline{4-4} \cline{5-5}
    & &
      \multirow{2}{*}{18 - 24 mo}
      &
        False & 77.28 \\
      & & &
        True & 22.72 \\ \cline{2-5}
  &
    \multirow{4}{*}{\begin{tabular}[c]{@{}l@{}}$EDSS_{mean}$\\ $ > 5$ (Severe\\ disability)\end{tabular}}
    &
      \multirow{2}{*}{0 - 6 mo}
      &
        False & 77.3 \\
      & & &
        True & 22.7 \\ \cline{3-3} \cline{4-4} \cline{5-5}
    & &
      \multirow{2}{*}{6 - 12 mo}
      &
        False & 79.8 \\
      & & &
        True & 20.2 \\ \cline{3-3} \cline{4-4} \cline{5-5}
    & &
      \multirow{2}{*}{12 - 18 mo}
      &
        False & 83.83 \\
      & & &
        True & 16.17 \\ \cline{3-3} \cline{4-4} \cline{5-5}
    & &
      \multirow{2}{*}{18 - 24 mo}
      &
        False & 87.73 \\
      & & &
        True & 12.27 \\ \cline{2-5}
  &
    \multirow{4}{*}{\begin{tabular}[c]{@{}l@{}}$EDSS_{mean}$\\ as severity\\ category\end{tabular}}
    &
      \multirow{2}{*}{0 - 6 mo}
      &
        No disability & 24.52 \\
      & & &
        Mild & 24.63 \\
      & & &
        Moderate & 26.17 \\
      & & &
        Severe & 24.66 \\ \cline{3-3} \cline{4-4} \cline{5-5}
    & &
      \multirow{2}{*}{6 - 12 mo}
      &
        No disability & 36.72 \\
      & & &
        Mild & 20.05 \\
      & & &
        Moderate & 21.57 \\
      & & &
        Severe & 21.65 \\ \cline{3-3} \cline{4-4} \cline{5-5}
    & &
      \multirow{2}{*}{12 - 18 mo}
      &
        No disability & 53.04 \\
      & & &
        Mild & 14.12 \\
      & & &
        Moderate & 15.47 \\
      & & &
        Severe & 17.35 \\ \cline{3-3} \cline{4-4} \cline{5-5}
    & &
      \multirow{2}{*}{18 - 24 mo}
      &
        No disability & 64.90 \\
      & & &
        Mild & 10.52 \\
      & & &
        Moderate & 11.42 \\
      & & &
        Severe & 13.14 \\ \hline
\end{tabular}%
}
\end{table*}

\section{Feature ablation studies}
\label{appendix:feature_ablation_studies}

Table \ref{feature_ablation_appendix} contains results for the feature ablation studies on both MSOAC and Floodlight. For MSOAC the label horizon chosen was 6-12 months, while for Floodlight it was 1-2 weeks.

\begin{table*}[]
\centering
\caption{Summary of feature ablation studies on MSOAC and Floodlight for 6-12 months and 1-2 weeks horizon respectively.}
\label{feature_ablation_appendix}
\resizebox{\textwidth}{!}{
\begin{tabular}{|c|l|c|cclcc|}
\hline
\multirow{2}{*}{\textbf{Dataset}} &
  \multicolumn{1}{c|}{\multirow{2}{*}{\textbf{\begin{tabular}[c]{@{}c@{}}Prediction\\ tasks\end{tabular}}}} &
  \multirow{2}{*}{\textbf{Models}} &
  \multicolumn{1}{c|}{\textbf{Feature Groups}} \\ \cline{4-8} 
 &
  \multicolumn{1}{c|}{} &
   &
  \multicolumn{1}{c|}{\textit{Demographics}} &
  \multicolumn{1}{c|}{\textit{\begin{tabular}[c]{@{}c@{}}Functional\\ Tests\end{tabular}}} &
  \multicolumn{1}{l|}{Questionnaires} &
  \multicolumn{1}{c|}{\textit{\begin{tabular}[c]{@{}c@{}}Performance\\ Outcome\\ Measures\end{tabular}}} &
  \textit{Full feature set} \\ \hline
\multirow{4}{*}{MSOAC} &
  \multirow{3}{*}{$EDSS_{mean}$} &
  Linear Regression &
  \multicolumn{1}{c|}{2.497 (0.091)} &
  \multicolumn{1}{c|}{2.138 (0.050)} &
  \multicolumn{1}{l|}{2.271 (0.052)} &
  \multicolumn{1}{c|}{2.131 (0.083)} &
  2.114 (0.111) \\ \cline{3-8} 
 &
   &
  Gradient Boosted Regressor &
  \multicolumn{1}{c|}{2.342 (0.062)} &
  \multicolumn{1}{c|}{2.045 (0.058)} &
  \multicolumn{1}{l|}{2.32 (0.063)} &
  \multicolumn{1}{c|}{1.951 (0.088)} &
  1.901(0.057) \\ \cline{3-8} 
 &
   &
  TCN &
  \multicolumn{1}{c|}{1.957 (0.051)} &
  \multicolumn{1}{c|}{1.777 (0.091)} &
  \multicolumn{1}{l|}{1.860 (0.049)} &
  \multicolumn{1}{c|}{1.676 (0.077)} &
  1.650 (0.067) \\ \cline{2-8} 
 &
  \multirow{4}{*}{\shortstack{\\$EDSS_{mean}$ \textgreater 3 \\ (Moderate disability)}} &
  Logistic Regression &
  \multicolumn{1}{c|}{0.529 (0.022)} &
  \multicolumn{1}{c|}{0.597 (0.025)} &
  \multicolumn{1}{l|}{0.631 (0.014)} &
  \multicolumn{1}{c|}{0.693 (0.011)} &
  0.707 (0.014) \\ \cline{3-8} 
 &
   &
  MLP &
  \multicolumn{1}{c|}{0.558 (0.020)} &
  \multicolumn{1}{c|}{0.662 (0.025)} &
  \multicolumn{1}{l|}{0.630 (0.027)} &
  \multicolumn{1}{c|}{0.716 (0.017)} &
  0.731 (0.019) \\ \cline{3-8} 
 &
   &
  Gradient Boosted Classifier &
  \multicolumn{1}{c|}{0.573 (0.020)} &
  \multicolumn{1}{c|}{0.67 (0.018)} &
  \multicolumn{1}{l|}{0.713 (0.015)} &
  \multicolumn{1}{c|}{0.75 (0.016)} &
  0.756 (0.019) \\ \cline{3-8} 
 &
   &
  TCN &
  \multicolumn{1}{c|}{0.678 (0.019)} &
  \multicolumn{1}{c|}{0.766 (0.025)} &
  \multicolumn{1}{l|}{0.789 (0.020)} &
  \multicolumn{1}{c|}{0.816 (0.037)} &
  0.82 (0.027) \\ \cline{2-8} 
 &
  \multirow{4}{*}{\shortstack{\\$EDSS_{mean}$ \textgreater 5 \\ (Severe disability)}} &
  Logistic Regression &
  \multicolumn{1}{c|}{0.344 (0.036)} &
  \multicolumn{1}{c|}{0.467 (0.032)} &
  \multicolumn{1}{l|}{0.491 (0.032)} &
  \multicolumn{1}{c|}{0.549 (0.031)} &
  0.576 (0.028) \\ \cline{3-8} 
 &
   &
  MLP &
  \multicolumn{1}{c|}{0.368 (0.038)} &
  \multicolumn{1}{c|}{0.557 (0.023)} &
  \multicolumn{1}{l|}{0.440 (0.038)} &
  \multicolumn{1}{c|}{0.556 (0.04)} &
  0.597 (0.027) \\ \cline{3-8} 
 &
   &
  Gradient Boosted Classifier &
  \multicolumn{1}{c|}{0.425 (0.04)} &
  \multicolumn{1}{c|}{0.570 (0.098)} &
  \multicolumn{1}{l|}{0.553 (0.012)} &
  \multicolumn{1}{c|}{0.641 (0.032)} &
  0.676 (0.032) \\ \cline{3-8} 
 &
   &
  TCN &
  \multicolumn{1}{c|}{0.456 (0.028)} &
  \multicolumn{1}{c|}{0.665 (0.036)} &
  \multicolumn{1}{l|}{0.608 (0.037)} &
  \multicolumn{1}{c|}{0.691 (0.034)} &
  0.722 (0.039) \\ \cline{2-8} 
 &
  \multirow{4}{*}{\begin{tabular}[c]{@{}l@{}}$EDSS_{mean}$ as\\ severity\\ category\end{tabular}} &
  Logistic Regression &
  \multicolumn{1}{c|}{0.302 (0.009)} &
  \multicolumn{1}{c|}{0.384 (0.017)} &
  \multicolumn{1}{l|}{0.423 (0.014)} &
  \multicolumn{1}{c|}{0.457 (0.014)} &
  0.470 (0.015) \\ \cline{3-8} 
 &
   &
  MLP &
  \multicolumn{1}{c|}{0.397 (0.013)} &
  \multicolumn{1}{c|}{0.521 (0.016)} &
  \multicolumn{1}{l|}{0.475 (0.019)} &
  \multicolumn{1}{c|}{0.582 (0.014)} &
  0.633 (0.018) \\ \cline{3-8} 
 &
   &
  Gradient Boosted Classifier &
  \multicolumn{1}{c|}{0.400 (0.010)} &
  \multicolumn{1}{c|}{0.515 (0.015)} &
  \multicolumn{1}{l|}{0.561 (0.031)} &
  \multicolumn{1}{c|}{0.630 (0.013)} &
  0.675 (0.016) \\ \cline{3-8} 
 &
   &
  TCN &
  \multicolumn{1}{c|}{0.520 (0.011)} &
  \multicolumn{1}{c|}{0.572 (0.086)} &
  \multicolumn{1}{l|}{0.659 (0.016)} &
  \multicolumn{1}{c|}{0.686 (0.042)} &
  0.709 (0.044) \\ \hline
\multirow{4}{*}{Floodlight} &
  \multirow{3}{*}{\begin{tabular}[c]{@{}l@{}}Cognitive\\ disability\\ score\end{tabular}} &
  Linear Regression &
  \multicolumn{1}{c|}{0.308 (0.018)} &
  \multicolumn{1}{c|}{0.305 (0.008)} &
  \multicolumn{1}{l|}{0.322 (0.021)} &
  \multicolumn{1}{c|}{0.303 (0.016)} &
  0.285 (0.015) \\ \cline{3-8} 
 &
   &
  Gradient Boosted Regressor &
  \multicolumn{1}{c|}{0.306 (0.017)} &
  \multicolumn{1}{c|}{0.286 (0.012)} &
  \multicolumn{1}{l|}{0.312 (0.019)} &
  \multicolumn{1}{c|}{0.283 (0.014)} &
  0.275 (0.014) \\ \cline{3-8} 
 &
   &
  TCN &
  \multicolumn{1}{c|}{0.306 (0.017)} &
  \multicolumn{1}{c|}{0.286 (0.012)} &
  \multicolumn{1}{l|}{0.416 (0.037)} &
  \multicolumn{1}{c|}{0.283 (0.014)} &
  0.275 (0.014)
   \\ \cline{2-8} 
 &
  \multirow{3}{*}{\begin{tabular}[c]{@{}l@{}}Dexterity\\ disability\\ score\end{tabular}} &
  Linear Regression &
  \multicolumn{1}{c|}{0.161 (0.015)} &
  \multicolumn{1}{c|}{0.163 (0.011)} &
  \multicolumn{1}{l|}{0.165 (0.012)} &
  \multicolumn{1}{c|}{0.161 (0.012)} &
  0.153 (0.012) \\ \cline{3-8} 
 &
   &
  Gradient Boosted Regressor &
  \multicolumn{1}{c|}{0.159 (0.014)} &
  \multicolumn{1}{c|}{0.153 (0.012)} &
  \multicolumn{1}{l|}{0.163 (0.013)} &
  \multicolumn{1}{c|}{0.152 (0.012)} &
  0.148 (0.012) \\ \cline{3-8} 
 &
   &
  TCN &
  \multicolumn{1}{c|}{0.159 (0.014)} &
  \multicolumn{1}{c|}{0.153 (0.012)} &
  \multicolumn{1}{l|}{0.198 (0.021)} &
  \multicolumn{1}{c|}{0.152 (0.012)} &
  0.148 (0.012)
   \\ \cline{2-8} 
 &
  \multirow{3}{*}{\begin{tabular}[c]{@{}l@{}}Mobility\\ disability\\ score\end{tabular}} &
  Linear Regression &
  \multicolumn{1}{l|}{0.283 (0.016)} &
  \multicolumn{1}{l|}{0.269 (0.017)} &
  \multicolumn{1}{l|}{0.298 (0.019)} &
  \multicolumn{1}{l|}{0.268 (0.021)} &
  \multicolumn{1}{l|}{0.256 (0.018)} \\ \cline{3-8} 
 &
   &
  Gradient Boosted Regressor &
  \multicolumn{1}{l|}{0.278 (0.018)} &
  \multicolumn{1}{l|}{0.249 (0.017)} &
  \multicolumn{1}{l|}{0.292 (0.018)} &
  \multicolumn{1}{l|}{0.248 (0.019)} &
  \multicolumn{1}{l|}{0.244 (0.021)} \\ \cline{3-8} 
 &
   &
  TCN &
  \multicolumn{1}{l|}{0.278 (0.018)} &
  \multicolumn{1}{l|}{0.249 (0.017)} &
  \multicolumn{1}{l|}{0.381 (0.031)} &
  \multicolumn{1}{l|}{0.248 (0.019)} &
  \multicolumn{1}{l|}{0.244 (0.021)} \\ \cline{2-8} 
 &
  \multirow{3}{*}{\begin{tabular}[c]{@{}l@{}}Overall\\ disability\\ score\end{tabular}} &
  Linear Regression &
  \multicolumn{1}{l|}{0.224 (0.014)} &
  \multicolumn{1}{l|}{0.222 (0.008)} &
  \multicolumn{1}{l|}{0.237 (0.017)} &
  \multicolumn{1}{l|}{0.220 (0.014)} &
  \multicolumn{1}{l|}{0.206 (0.012)} \\ \cline{3-8} 
 &
   &
  \multicolumn{1}{l|}{Gradient Boosted Regressor} &
  \multicolumn{1}{l|}{0.220 (0.012)} &
  \multicolumn{1}{l|}{0.206 (0.012)} &
  \multicolumn{1}{l|}{0.230 (0.016)} &
  \multicolumn{1}{l|}{0.205 (0.013)} &
  \multicolumn{1}{l|}{0.197 (0.013)} \\ \cline{3-8} 
 &
   &
  TCN &
  \multicolumn{1}{l|}{0.220 (0.018)} &
  \multicolumn{1}{l|}{0.206 (0.012)} &
  \multicolumn{1}{l|}{0.308 (0.034)} &
  \multicolumn{1}{l|}{0.205 (0.013)} &
  \multicolumn{1}{l|}{0.197 (0.013)} \\ \hline
\end{tabular}
}
\end{table*}

\section{Subgroup analysis}
\label{appendix:subgroup_results}

Table \ref{msoac_subgroup_results} presents subgroup results for the classification tasks performed on the MSOAC dataset, for the 6-12 month horizon.

\begin{table*}[]
\centering
\caption{Subgroup results for prediction tasks in MSOAC on the 6-12 months horizon.}
\label{msoac_subgroup_results}
\resizebox{1\textwidth}{!}{%
\begin{tabular}{|c|c|c|c|c|c|c|c|}
\hline
\textbf{Tasks} &
  \textbf{Models} &
  \textbf{Female} &
  \textbf{Male} &
  \textbf{Age \textless{30}} &
  \textbf{Age 30-50} &
  \textbf{Age 50-70} &
  \textbf{Age \textgreater 70}\\ \hline
\multirow{1}{*}{Instance count} &
    \multirow{1}{*}{-} &
      23028 &
      11604 &
      3643 &
      17151 &
      7489 &
      11 \\ \hline
\multirow{4}{*}{\textit{$EDSS_{mean}$ \textgreater 3}} &
  \multirow{1}{*}{Logistic Regression} &
      \begin{tabular}[c]{@{}l@{}}0.71\\ (0.027)\end{tabular} &
      \begin{tabular}[c]{@{}l@{}}0.72\\ (0.016)\end{tabular} &
      \begin{tabular}[c]{@{}l@{}}0.63\\ (0.055)\end{tabular} &
      \begin{tabular}[c]{@{}l@{}}0.65\\ (0.017)\end{tabular} &
      \begin{tabular}[c]{@{}l@{}}0.77\\ (0.022)\end{tabular} &
      \begin{tabular}[c]{@{}l@{}}1.0\\ (0.0)\end{tabular} \\ \cline{2-8} 
 & \multirow{1}{*}{MLP} &
      \begin{tabular}[c]{@{}l@{}}0.74\\ (0.024)\end{tabular} &
      \begin{tabular}[c]{@{}l@{}}0.72\\ (0.027)\end{tabular} &
      \begin{tabular}[c]{@{}l@{}}0.70\\ (0.090)\end{tabular} &
      \begin{tabular}[c]{@{}l@{}}0.67\\ (0.029)\end{tabular} &
      \begin{tabular}[c]{@{}l@{}}0.71\\ (0.032)\end{tabular} &
      \begin{tabular}[c]{@{}l@{}}0.90\\ (0.0)\end{tabular} \\ \cline{2-8} 
 &
  \multirow{1}{*}{\begin{tabular}[c]{@{}c@{}}Gradient Boosted Classifier\end{tabular}} &
      \begin{tabular}[c]{@{}l@{}}0.76\\ (0.029)\end{tabular} &
      \begin{tabular}[c]{@{}l@{}}0.75\\ (0.025)\end{tabular} &
      \begin{tabular}[c]{@{}l@{}}0.76\\ (0.068)\end{tabular} &
      \begin{tabular}[c]{@{}l@{}}0.71\\ (0.023)\end{tabular} &
      \begin{tabular}[c]{@{}l@{}}0.78\\ (0.031)\end{tabular} &
      \begin{tabular}[c]{@{}l@{}}1.0\\ (0.0)\end{tabular} \\ \cline{2-8} 
 &
  \multirow{1}{*}{TCN} &
      \begin{tabular}[c]{@{}l@{}}0.81\\ (0.032)\end{tabular} &
      \begin{tabular}[c]{@{}l@{}}0.84\\ (0.034)\end{tabular} &
      \begin{tabular}[c]{@{}l@{}}0.61\\ (0.171)\end{tabular} &
      \begin{tabular}[c]{@{}l@{}}0.78\\ (0.048)\end{tabular} &
      \begin{tabular}[c]{@{}l@{}}0.88\\ (0.033)\end{tabular} &
      \begin{tabular}[c]{@{}l@{}}0.81\\ (0.0)\end{tabular} \\ \hline
\multirow{4}{*}{\textit{$EDSS_{mean}$  \textgreater 5}} &
  \multirow{1}{*}{Logistic Regression} &
      \begin{tabular}[c]{@{}l@{}}0.56\\ (0.035)\end{tabular} &
      \begin{tabular}[c]{@{}l@{}}0.60\\ (0.024)\end{tabular} &
      \begin{tabular}[c]{@{}l@{}}0.41\\ (0.095)\end{tabular} &
      \begin{tabular}[c]{@{}l@{}}0.52\\ (0.043)\end{tabular} &
      \begin{tabular}[c]{@{}l@{}}0.63\\ (0.045)\end{tabular} &
      \begin{tabular}[c]{@{}l@{}}NaN\\ (NaN)\end{tabular} \\ \cline{2-8} 
 & \multirow{1}{*}{MLP} &
      \begin{tabular}[c]{@{}l@{}}0.60\\ (0.053)\end{tabular} &
      \begin{tabular}[c]{@{}l@{}}0.58\\ (0.040)\end{tabular} &
      \begin{tabular}[c]{@{}l@{}}0.57\\ (0.0125)\end{tabular} &
      \begin{tabular}[c]{@{}l@{}}0.54\\ (0.028)\end{tabular} &
      \begin{tabular}[c]{@{}l@{}}0.55\\ (0.049)\end{tabular} &
      \begin{tabular}[c]{@{}l@{}}NaN\\ (NaN)\end{tabular} \\ \cline{2-8} 
 &
  \multirow{1}{*}{\begin{tabular}[c]{@{}c@{}}Gradient Boosted Classifier\end{tabular}} &
      \begin{tabular}[c]{@{}l@{}}0.68\\ (0.043)\end{tabular} &
      \begin{tabular}[c]{@{}l@{}}0.67\\ (0.058)\end{tabular} &
      \begin{tabular}[c]{@{}l@{}}0.66\\ (0.116)\end{tabular} &
      \begin{tabular}[c]{@{}l@{}}0.64\\ (0.042)\end{tabular} &
      \begin{tabular}[c]{@{}l@{}}0.71\\ (0.045)\end{tabular} &
      \begin{tabular}[c]{@{}l@{}}NaN\\ (NaN)\end{tabular} \\ \cline{2-8} 
 &
  \multirow{1}{*}{TCN} &
      \begin{tabular}[c]{@{}l@{}}0.72\\ (0.061)\end{tabular} &
      \begin{tabular}[c]{@{}l@{}}0.72\\ (0.041)\end{tabular} &
      \begin{tabular}[c]{@{}l@{}}0.43\\ (0.324)\end{tabular} &
      \begin{tabular}[c]{@{}l@{}}0.68\\ (0.057)\end{tabular} &
      \begin{tabular}[c]{@{}l@{}}0.78\\ (0.051)\end{tabular} &
      \begin{tabular}[c]{@{}l@{}}0.58\\ (0.0)\end{tabular} \\ \hline
\multirow{4}{*}{\begin{tabular}[c]{@{}l@{}}$EDSS_{mean}$\\ as severity\\ category\end{tabular}} &
  \multirow{1}{*}{Logistic Regression} &
      \begin{tabular}[c]{@{}l@{}}0.47\\ (0.018)\end{tabular} &
      \begin{tabular}[c]{@{}l@{}}0.48\\ (0.023)\end{tabular} &
      \begin{tabular}[c]{@{}l@{}}0.48\\ (0.033)\end{tabular} &
      \begin{tabular}[c]{@{}l@{}}0.48\\ (0.027)\end{tabular} &
      \begin{tabular}[c]{@{}l@{}}0.48\\ (0.028)\end{tabular} &
      \begin{tabular}[c]{@{}l@{}}0.70\\ (0.0)\end{tabular} \\ \cline{2-8} 
 & \multirow{1}{*}{MLP} &
      \begin{tabular}[c]{@{}l@{}}0.63\\ (0.019)\end{tabular} &
      \begin{tabular}[c]{@{}l@{}}0.63\\ (0.028)\end{tabular} &
      \begin{tabular}[c]{@{}l@{}}0.61\\ (0.045)\end{tabular} &
      \begin{tabular}[c]{@{}l@{}}0.60\\ (0.019)\end{tabular} &
      \begin{tabular}[c]{@{}l@{}}0.58\\ (0.028)\end{tabular} &
      \begin{tabular}[c]{@{}l@{}}0.67\\ (0.0)\end{tabular} \\ \cline{2-8} 
 &
  \multirow{1}{*}{\begin{tabular}[c]{@{}c@{}}Gradient Boosted\\ Classifier\end{tabular}} &
      \begin{tabular}[c]{@{}l@{}}0\\ (0.0)\end{tabular} &
      \begin{tabular}[c]{@{}l@{}}0\\ (0.0)\end{tabular} &
      \begin{tabular}[c]{@{}l@{}}0.69\\ (0.039)\end{tabular} &
      \begin{tabular}[c]{@{}l@{}}0.60\\ (0.026)\end{tabular} &
      \begin{tabular}[c]{@{}l@{}}0.62\\ (0.031)\end{tabular} &
      \begin{tabular}[c]{@{}l@{}}0.62\\ (0.0)\end{tabular} \\ \cline{2-8} 
 &
  \multirow{1}{*}{TCN} &
      \begin{tabular}[c]{@{}l@{}}0.71\\ (0.050)\end{tabular} &
      \begin{tabular}[c]{@{}l@{}}0.71\\ (0.041)\end{tabular} &
      \begin{tabular}[c]{@{}l@{}}0.57\\ (0.085)\end{tabular} &
      \begin{tabular}[c]{@{}l@{}}0.62\\ (0.056)\end{tabular} &
      \begin{tabular}[c]{@{}l@{}}0.60\\ (0.065)\end{tabular} &
      \begin{tabular}[c]{@{}l@{}}0.52\\ (0.0)\end{tabular} \\ \hline
\end{tabular}
}
\end{table*}

\section{Ethical considerations and broader impact}
\label{ethical_consideration}
Employing easily accessible information in diagnosis and predicting the progression of MS, can have many advantages, including but not limited to better choices of treatment and interventions for each patient, and hopefully reducing the number of relapses in RRMS and hence, the disability of patients. 
However, these studies should be done with a great amount of care. First, multiple studies have shown great disparity of results between various demographics, typically stemming from representation issues in datasets. Besides potential disparity of results among demographics, we should pay great care about where and how this research is being used. This is of importance, especially where the resources are scarce. In addition, it is important to note, we intend the outcome of this study to be used for patients' access to better choices of treatment and not for this information to be used for unintended purposes such as insurance policies.

\end{document}